\documentclass[lettersize,journal]{IEEEtran}
\usepackage{amsmath,amsfonts}
\usepackage{algorithmic}
\usepackage{algorithm}
\usepackage{array}
\usepackage[caption=false,font=normalsize,labelfont=sf,textfont=sf]{subfig}
\usepackage{textcomp}
\usepackage{stfloats}
\usepackage{url}
\usepackage{verbatim}
\usepackage{graphicx}
\usepackage{cite}
\usepackage{color}
\usepackage[colorlinks,bookmarks=false,citecolor=green]{hyperref}

\usepackage{booktabs}
\usepackage{multirow}

\usepackage{pifont}
\newcommand{\cmark}{\ding{51}}
\newcommand{\xmark}{\ding{55}}

\begin{document}

\newcommand{\Method}{SkeleMixCLR}
\newcommand{\Methodplus}{SkeleMixCLR+}
\newcommand{\aug}{SkeleMix}
\newcommand{\trimmed}{trimmed}
\newcommand{\truncated}{truncated}

\title{Contrastive Learning from Spatio-Temporal Mixed Skeleton Sequences for Self-Supervised Skeleton-Based Action Recognition}

\author{Zhan Chen, Hong Liu$^{\dag}$, Tianyu Guo, Zhengyan Chen, Pinhao Song, and Hao Tang
\thanks{$\dag$: corresponding author}
\thanks{This work is supported by National Key R\&D Program of China (No. 2020AAA0108904), Science and Technology Plan of Shenzhen (No. JCYJ20200109140410340)}
\thanks{Z. Chen, H. Liu, T. Guo, Z. Chen and P. Song are with the Key
	Laboratory of Machine Perception, Peking University, Shenzhen Graduate
	School, Beijing 100871, China (e-mail: zhanchen\_cz@pku.edu.cn; hongliu@pku.edu.cn; levigty@stu.pku.edu.cn, chenzhengyan@pku.edu.cn, pinhaosong@pku.edu.cn)}
\thanks{Hao Tang is with the Department of Information Technology and Electrical Engineering, ETH Zurich,  Zurich 8092, Switzerland (e-mail: hao.tang@vision.ee.ethz.ch)}
}

\markboth{IEEE TRANSACTIONS ON MULTIMEDIA}%
{Chen \MakeLowercase{\textit{et al.}}: Contrastive Learning from Spatio-Temporal Mixed Skeleton Sequences for Self-Supervised Skeleton-Based Action Recognition}


\maketitle

\begin{abstract}
Self-supervised skeleton-based action recognition with contrastive learning has attracted much attention.
%
%
Recent literature shows that data augmentation and large sets of contrastive pairs are crucial in learning such representations.
In this paper, we found that directly extending contrastive pairs based on normal augmentations brings limited returns in terms of performance, because the contribution of contrastive pairs from the normal data augmentation to the loss get smaller as training progresses.
Therefore, we delve into hard contrastive pairs for contrastive learning.
Motivated by the success of mixing augmentation strategy which improves the performance of many tasks by synthesizing novel samples,
we propose \Method: a contrastive learning framework with a spatio-temporal skeleton mixing augmentation (\aug) to complement current contrastive learning approaches by providing hard contrastive samples.
First, \aug~utilizes the topological information of skeleton data to mix two skeleton sequences by randomly combing the cropped skeleton
fragments (the \trimmed~view) with the remaining skeleton
sequences (the \truncated~view).
Second, a spatio-temporal mask pooling is applied to separate these two views at the feature level.
Third, we extend contrastive pairs with these two views.
\Method~leverages the \trimmed~and \truncated~views to provide abundant hard contrastive pairs since they involve some context information from each other due to the graph convolution operations, which allows the model to learn better motion representations for action recognition. 
Extensive experiments on NTU-RGB+D, NTU120-RGB+D, and PKU-MMD datasets show that \Method~achieves state-of-the-art performance.
Codes are available at \url{https://github.com/czhaneva/SkeleMixCLR}.

\end{abstract}

\begin{IEEEkeywords}
Self-supervised learning, contrastive learning, hard contrastive pairs, skeleton-based action recognition.
\end{IEEEkeywords}

\section{Introduction}
    \IEEEPARstart{A}{s} a critical problem in computer vision, human action recognition has been researched for decades, since this task enjoys various applications, such as video surveillance, human-machine interaction, virtual reality, video analysis, and so on \cite{aggarwal2011human,poppe2010survey,sudha2017approaches,weinland2011survey,sun2022human}.
    With the fast development of the depth sensors \cite{smisek20133d} and advanced pose estimation algorithm \cite{li2022mhformer,strided,hua2022weakly}, 3D skeleton data becomes more accessible, making skeleton-based action recognition an essential branch in studying human action dynamics.
    In the past decades, skeleton-based action recognition has evolved by leaps and bounds, and many promising methods have emerged \cite{zhang2018fusing,yan2018spatial,li2019actional,shi2019two,liu2020disentangling,cheng2020skeleton,zhang2020deep}.
    However, most of these methods follow a fully supervised framework, which requires numerous expensive annotated data.
    Therefore, self-supervised skeleton-based action recognition, which utilizes large amounts of unlabeled data to guide models to learn discriminative spatio-temporal motion representations, has emerged as a new popular research branch.

	\begin{figure}[t]
		\centering
		\includegraphics[width=1\linewidth]{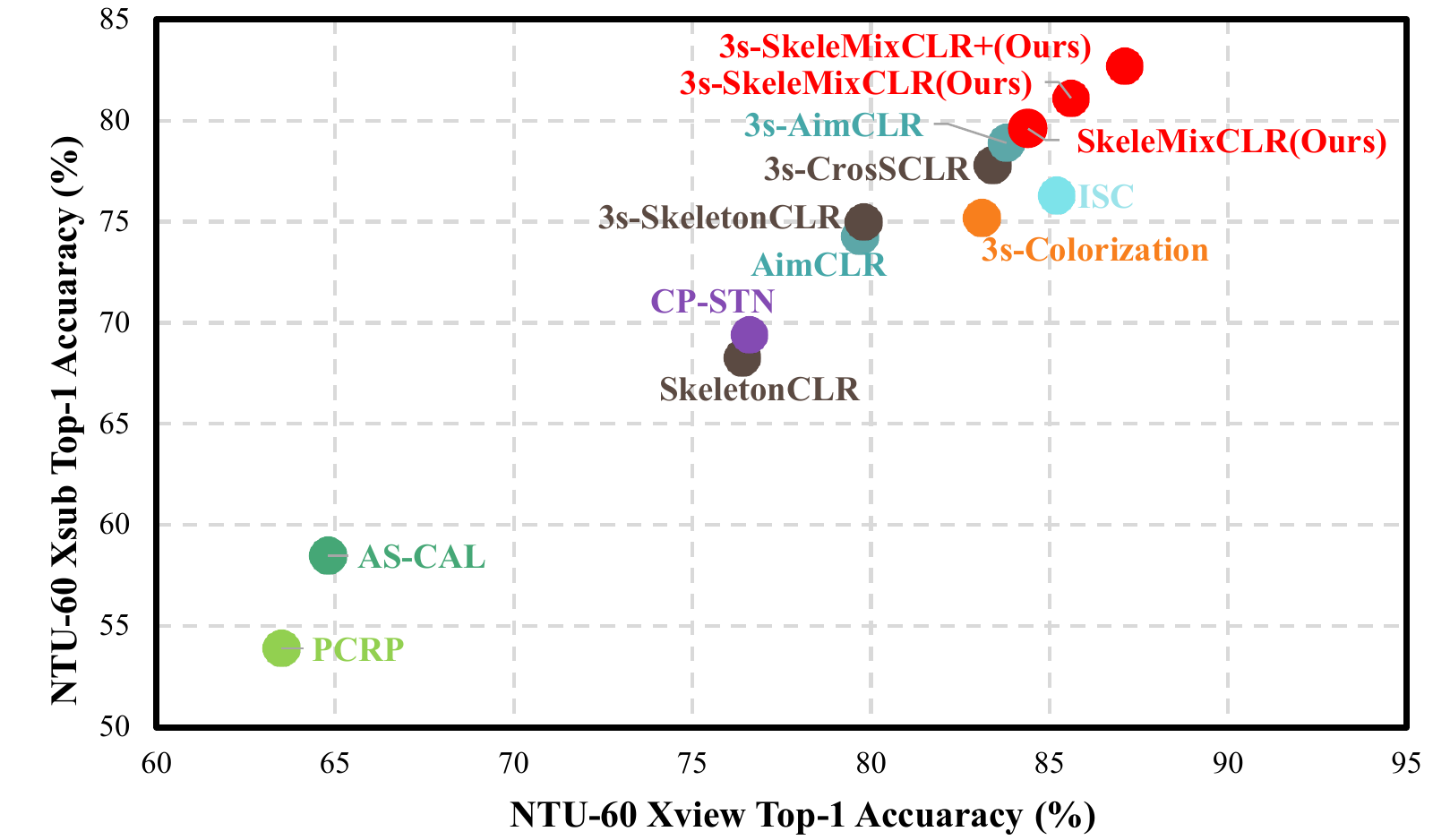} 
		\caption{Comparison of various methods on NTU-RGB+D cross-subject (Xsub) and cross-view (Xview) benchmarks. Our method achieves state-of-the-art performance.}
		\label{fig:performance}
	\end{figure}
	
	Several self-supervised works focus on designing pretext tasks, such as sequence reconstruction \cite{zheng2018unsupervised}, jigsaw puzzle \cite{lin2020ms2l}, or motion prediction \cite{cheng2021motion}, to help the model learn generalized features.
	Nevertheless, the joint-level pretext tasks require the model to learn fine-grained features invariant to viewpoint changes and skeleton scale, rather than focusing on higher-level semantic features relevant to skeleton-based action recognition.
	Recently, contrastive learning has become a key component of component self-supervised skeleton-based action recognition \cite{rao2021augmented,guo2021contrastive,li20213d}.
	Contrastive learning based methods typically apply data augmentation on the skeleton sequences to generate different views of sequences and construct contrastive pairs between different views, then guide the model to learn spatio-temporal representations by pulling the positive pairs closer and pushing the negative samples away using a contrastive loss.
	Compared with the joint-level pretext tasks, contrastive learning based methods focus more on high-level context information, making the learned representations better for the downstream tasks.
	
    The literature shows that augmentation and large sets of contrastive pairs play crucial roles in learning such representations \cite{he2020momentum,chen2020simple,kalantidis2020hard}.
    Based on this, we try to improve contrastive learning by providing more contrastive pairs.
    However, as found in our experiments, naively extending contrastive pairs based on normal augmentations brings limited returns in terms of performance, because the contrastive pairs from the normal data augmentation contribute less and less to the loss as training progresses.

    Considering that the contrastive pairs from the normal data augmentation provide insufficient information, which limits the ability to explore novel movement patterns, we therefore delve into hard contrastive pairs.
    At the same time, mixing augmentation strategy improves performance of many top-performing methods in various tasks by synthesizing novel samples, forcing the model to learn generalized and robust features \cite{zhang2017mixup,yun2019cutmix,verma2019manifold,song2021achieving,jing2022adversarial}.
    Motivated by the above, we propose \Method: a contrastive learning framework with a spatio-temporal skeleton mixing augmentation (\aug) to complement current contrastive learning approaches by providing hard contrastive samples.
    First, we propose a unique skeleton sequence augmentation strategy called \aug~for contrastive learning.
	Specifically, skeleton joints are first partitioned into five parts based on the topological information of skeleton data, and several body parts are randomly selected for cropping.
    Then, the cropped skeleton fragments (the \trimmed~view) are randomly combined with the remaining skeleton sequences (called the \truncated~view) to mix skeleton sequences.
    %
	%
	Second, we use a spatio-temporal mask pooling (STMP) operation to separate the \trimmed~and \truncated~views and get the embeddings of the corresponding views.
	Since the \trimmed~and \truncated~views share some context information from each other, they provide hard samples for contrastive learning.
	Finally, we combine our method with the baseline method SkeletonCLR \cite{li20213d} as \Method, and follow the general approach to construct contrastive pairs, \emph{i.e.,} each view is positive with the views augmented from the same original skeleton sequences and negative with views augmented from other skeleton sequences \cite{he2020momentum,chen2020simple}.
	Moreover, we propose to perform multiple \aug~augmentation to provide diverse hard samples (\Methodplus), which can further boost the performance.
	Our method enables the model to learn better local and global spatio-temporal action representations by utilizing the mixed skeleton sequences to provide plenty of hard positive pairs such as the \truncated~view and the key view (as well as the \trimmed~view and the key view), and hard negative pairs such as the \trimmed~view and the \truncated~view, which helps to extract generalized representations for downstream tasks. 
	
	%
    Figure~\ref{fig:performance} shows the superiority of our proposed method.
    \Method~substantially improves the baseline SkeletonCLR by $8\% {\sim} 11.3\%$ on NTU-RGB+D benchmark \cite{shahroudy2016ntu}.
    The single stream \Method~outperforms many other methods, such as PCRP \cite{xu2021prototypical}, AS-CAL \cite{rao2021augmented}, CP-STN \cite{zhan2021spatial}, ISC \cite{thoker2021skeleton}, and even some ensemble methods, such as 3s-CrosSCLR \cite{li20213d}, 3s-AimCLR \cite{guo2021contrastive}, and 3s-Colorization \cite{yang2021skeleton}.
	Our contributions can be summarized as follows:
	\begin{itemize}
		\item We propose \aug~augmentation to provide abundant hard samples for self-supervised skeleton-based action recognition by combining the topological information of skeleton data.
		\item A simple yet effective framework \Method~based on \aug~is proposed.
		The \Method~facilitates the model to learn better discriminative global and local representations by introducing extensive hard contrastive pairs, which helps to achieve better performance in downstream tasks.
		\item A multiple \aug~augmentation strategy is proposed to provide diverse hard samples, which can further boost the performance.
		\item Extensive experimental results on NTU-RGB+D, NTU120-RGB+D, and PKU-MMD datasets show that the proposed \Method~achieves state-of-the-art performance under a variety of evaluation protocols.
	\end{itemize}
	
	\section{Related Work}
	\subsection{Self-Supervised Contrastive Learning}
	Contrastive learning, whose goal is increasing the similarity between positive pairs and decreasing the similarity between negative pairs, has shown promising performance in self-supervised representation learning \cite{he2020momentum,chen2020simple}.
	In the past few years, there emerged numerous self-supervised representation learning works based on contrastive learning, such as instance discrimination \cite{wu2018unsupervised}, SwAv \cite{caron2020unsupervised}, MoCo \cite{he2020momentum}, SimCLR \cite{chen2020simple}, BYOL \cite{grill2020bootstrap}, contrastive cluster \cite{li2021contrastive}, DINO \cite{caron2021emerging}, and SimSiam \cite{chen2021exploring}.
	These methods have achieved advanced results and are easy to transfer to other areas, such as skeleton-based action recognition.
	In this paper, we follow MoCov2 \cite{chen2020improved} framework to implement our method.
	
	\subsection{Mixing Augmentation Strategy}
	Most of the top-performing contrastive methods leverage data augmentations, which is crucial in learning useful representations because they modulate the hardness of the self-supervised task via the contrastive pairs.
	Recently, Mixing \cite{zhang2017mixup} and CutMix \cite{yun2019cutmix} are widely discussed in self-supervised contrastive learning.
	These operations are mainly performed between embedding features or between samples.
	MoCHi \cite{kalantidis2020hard} proposes a hard negative mixing strategy, which generates hard negative samples by mixing the embedding features to improve the generalization of the learned visual representations.
	Vi-Mix \cite{das2021vi} proposes CMMC to mix the data across different modalities of a video in their intermediate representations.
	MixCo \cite{kim2020mixco}, Un-Mix \cite{shen2020mix} and i-Mix \cite{lee2020mix} perform mixing operation between samples and generate corresponding smoothing labels to let the model be aware of the soft degree of similarity between contrastive pairs.
	The above methods mainly construct contrastive pairs with the mixing features.
	Different from these methods, we utilize the features of the \trimmed~and \truncated~views separated by STMP to provide hard samples for contrastive learning.
	Our method not only makes better use of the augmented features, but also improves the ability of the model to learn both local and global representations.
	RegionCL \cite{xu2021regioncl} also propose to utilize the separated features, while our method leverages the topological information of the skeleton data to maintain the consistency of action information, and performs mixing operation on both spatial and temporal domain.
	Moreover, we propose to apply multiple \aug~operation to provide a more adequate hard contrastive pairs.
	
	\subsection{Self-Supervised Skeleton-Based Action Recognition}
	Self-supervised skeleton-based action recognition has emerged as one of the promising direction for action recognition.
	LongT GAN \cite{zheng2018unsupervised} and P\&C \cite{su2020predict} rely on regeneration of the skeleton sequences to help the model to learn spatio-temporal representations.
	Colorization \cite{yang2021skeleton} leverages the colorized skeleton point cloud and designs an auto-encoder framework that can effectively learn spatio-temporal features from the artificial color labels of skeleton joints.
	With the development of contrastive learning, the past few years have witnessed a surge of successful self-supervised contrastive skeleton-based action recognition. 
	AS-CAL \cite{rao2021augmented} exploits eight different skeleton sequence augmentations and their combinations to generate query and key views for contrastive learning.
	ISC \cite{thoker2021skeleton} proposes inter-skeleton contrastive learning to enhance the learned features via different input skeleton representations.
	MS$^{2}$L \cite{lin2020ms2l} and CP-STN \cite{zhan2021spatial} combine contrastive learning with multi-pretext tasks such as masked sequences prediction, enabling the model to fully extract discriminative representations with spatio-temporal information.
	Works like CrosSCLR \cite{li20213d} and AimCLR \cite{guo2021contrastive} also try to improve contrastive learning by introducing extra hard contrastive pairs.
	In CrosSCLR, a cross-view knowledge mining strategy is developed to exam the similarity of samples, and select the most similar pairs as positive ones to boost the positive set in complementary views.
	In AimCLR, an extreme augmentation strategy is proposed to introduce movement patterns, which forces the model to learn more general representations by providing harder contrastive pairs.
    In this paper, we propose a unique skeleton sequence augmentation strategy \aug~to provide hard contrastive samples, which guides the model to learn better local and global spatio-temporal representations.
	
	\section{Methodology}
	\subsection{Overview of SkeletonCLR}\label{sec:skeletonclr}
\begin{figure*}[t]
		\centering
		\includegraphics[width=1\linewidth]{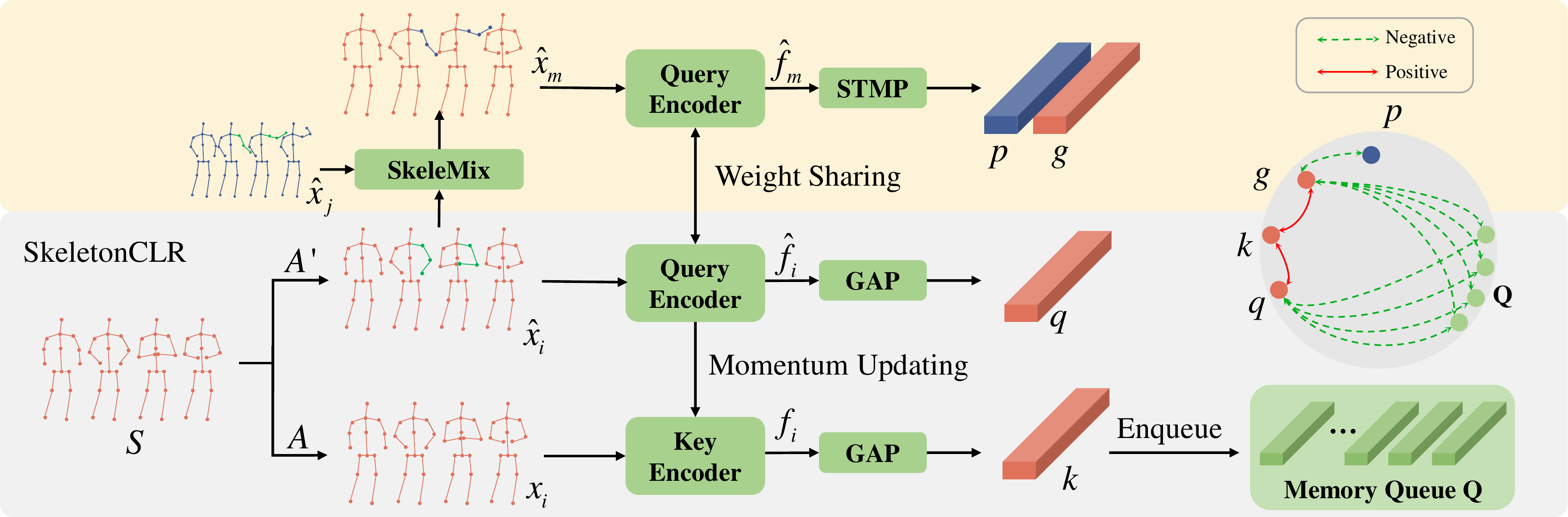} 
		\caption{Illustration of the proposed \Method. $S$ indicates the input skeleton sequence. 
				 Firstly, Two different data augmentations are applied to $S$ to get two different augmented skeleton sequences $x_{i}$ and $\hat{x}_{i}$, 
				 then \aug~described in Sec.~\ref{sec:skelemix} is adopted to generate the mixed skeleton sequence $x_{m}$ by performing mixing operation between $\hat{x}_{i}$ and $\hat{x}_{j}$, where $\hat{x}_{j}$ is another skeleton sequence in the same batch with $\hat{x_{i}}$.
				 Secondly, all skeleton sequences $x_{i}$, $\hat{x}_{i}$, $\hat{x}_{m}$ are fed into encoder to extract features, then a global average pooling operation (GAP) is applied to $\hat{f}_{i}$ and $f_{i}$ to get the query embedding $q$ and the key embedding $k$.  
				 The key embedding $k$ is stored in the memory queue \textbf{Q} to provide negative samples.
	 			 A STMP described in Sec.~\ref{sec:skelemixclr} is applied to separate embedding of the cropped skeleton fragment $p$ and embedding of the remaining skeleton sequence $g$ from the mixing feature $\hat{f}_{m}$.
				 Finally, $q$, $k$, $p$, $g$, and $\textbf{Q}$ are used to construct contrastive pairs.} 
		\label{fig:pipline}
	\end{figure*}

	As illustrated in the gray box of Figure~\ref{fig:pipline}, SkeletonCLR \cite{li20213d} follows the MoCov2 framework \cite{chen2020improved} to learn skeleton-based action representations.
	Given a skeleton sequence $S$, two different augmentations $A$ and $A'$ are applied to $S$ to obtain query sample $\hat{x}$ and key sample $x$.
	We denote $x,\hat{x} \in \mathbb{R}^{C \times T \times V}$, where $C$, $T$, and $V$ are the number of channels, frames, and nodes, respectively. 
	A query encoder and a momentum updated key encoder followed by global average pooling (GAP) are used to get query embedding $q$ and key embedding $k$. 
	Then the key embedding $k$ is stored in a first-in-first-out memory queue $\textbf{Q}=\{m_{i}\}^{K}_{i=1}$, where $K$ denotes the queue size.
	Following the criteria for constructing contrastive pairs in MoCov2, $q$ and $k$ form positive pairs while $q$ and the embeddings in $\textbf{Q}$ form negative pairs.
	The InfoNCE loss \cite{oord2018representation} formulated as Eq. \eqref{eq:infonce} is used to train the network, where $\cdot$ is dot product to compute the similarity between two $L2$ normalized embeddings, and $\tau$ is the temperature hyperparameter (set to 0.2 by default). 
	$Z(v)=\sum_{i=1}^{K}\exp(v \cdot m_{i} / \tau)$ denotes the similarity between embeddings of view $v$ and memory queue \textbf{Q}.
	\begin{equation}\label{eq:infonce}
	\mathcal{L}_{Info} = -\log(\frac{\exp(q \cdot k / \tau)}{\exp(q \cdot k / \tau) + Z(q)}).
	\end{equation}
	
	The parameters of query encoder $\theta_{q}$ are updated via gradient back propagation while the parameters of key encoder $\theta_{k}$ are updated as a moving-average of the query encoder, which can be formulated as:
	\begin{equation}\label{mom}
	\theta_{k} \leftarrow m\theta_{k} + (1-m)\theta_{q},
	\end{equation}
	where $m \in [0,1)$ is a momentum coefficient and typically close to 1 to maintain consistency of the embeddings in the memory queue.

	\subsection{\aug~Augmentation}\label{sec:skelemix}
	\begin{figure}[t]
		\centering
		\includegraphics[width=1\linewidth]{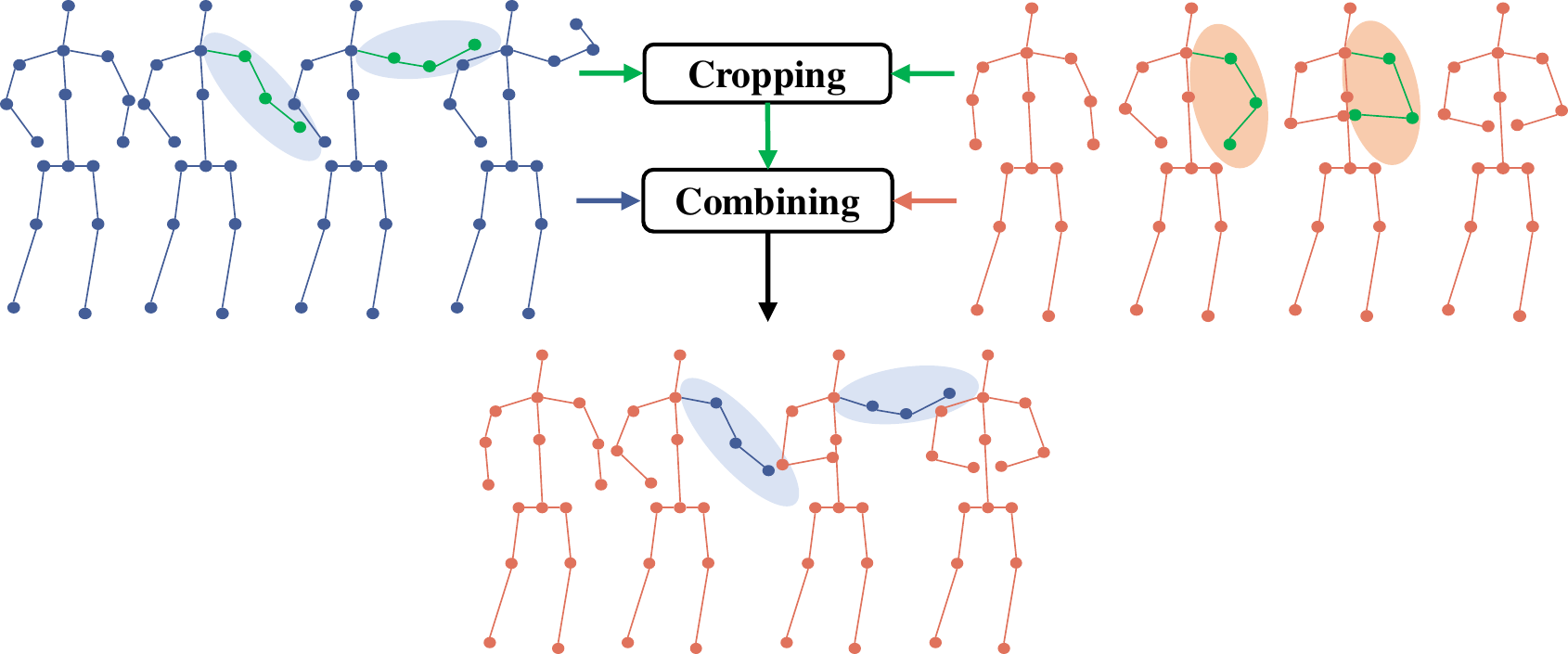} 
		\caption{Illustration of \aug~augmentation. The left hands (green) of instance waving (blue) and instance clapping (orange) in the second and third frames are cropped. Then, the left hand of instance waving is combined with the remaining skeleton sequence of clapping instance to generate the mixed skeleton sequence.} 
		\label{fig:aug}
	\end{figure}
	
	In this section, we will introduce our \aug~augmentation, which is a spatio-temporal mixing augmentation strategy.
	As illustrated in Figure~\ref{fig:aug}, our \aug~follows CutMix to peroform spatio-temporal mixing on skeleton sequences.
	Specifically, in the cropping step, skeleton joints are first partitioned into five part subsets $\mathcal{P}=\{$\emph{left-hand}, \emph{right-hand}, \emph{left-leg}, \emph{right-leg}, \emph{trunk}$\}$ based on the topological information of skeleton data.
	Second, two discrete uniform distributions $\mathcal{B}_{s} {\sim} U(\mathcal{B}_{sl}, \mathcal{B}_{su})$ and  $\mathcal{B}_{t} {\sim} U(\mathcal{B}_{tl}, \mathcal{B}_{tu})$ are used to determine the number of body parts to crop out and the duration of the cropped skeleton fragments.
	We sample once in each training iteration and obtain $\mathcal{N}_{s}$ and $\mathcal{N}_{t}$. 
	Then, we randomly choose $\mathcal{N}_{s}$ body parts from $\mathcal{P}$ to get the cropped skeleton joints $\mathcal{S}$, and randomly sample the start frame $t_s$ from a valid range that guarantees the completeness of the cropped skeleton fragments.
	Combining $\mathcal{N}_{t}$ and $t_{s}$, we can obtain the corresponding temporal cropping region $\mathcal{T}$. 
	Finally, we perform spatio-temporal cropping operation on $\hat{x}$ to get the cropped skeleton fragments.
	It is notable that the cropped regions ($\mathcal{S}$ and $\mathcal{T}$) within a mini-batch of the training skeleton sequences are typically the same to maintain consistency of action information.
	In combining step, we randomly combine the cropped skeleton fragments (named as the \trimmed~view) and the remaining skeleton sequences (named as the \truncated~view) to generate the mixed skeleton sequences.
	Thus, the mixed skeleton sequences consist of the \trimmed~view and \truncated~view.
	Furthermore, to provide a sufficient hard samples for contrastive learning, we propose to perform multiple \aug~augmentation.
	For convenience, we denote the mixed skeleton sequences generated by the $r_{th}$ \aug~augmentation as $\hat{x}^{r}_{m}, r\in(1,...,R)$, where $R$ denotes the total number of \aug~augmentation performed.
    \\
	\subsection{\Method}\label{sec:skelemixclr}
	In this section, we introduce our \Method~in detail.
	A weight-sharing encoder with the query view in SkeletonCLR \cite{li20213d} is used to extract features $\hat{f}^{r}_{m}$ from the mixed skeleton sequences $\hat{x}^{r}_{m}$.
	We denote $\hat{f}^{r}_{m} \in \mathbb{R}^{C' \times T' \times V}$, where $C'$ is the feature dimension and $T' = T/\mathcal{R}$ where $\mathcal{R}$ is the temporal downsampling ratio of the model.  
	To separate the embedding of the \trimmed~view $p^{r}$ and the \truncated~view $g^{r}$, we utilize a spatio-temporal mask pooling operation (STMP), which can be formulated as:
	\begin{equation}
	\begin{aligned}
	p^{r} &= (\sum^{T'}\sum^{V}f^{r}_{m} \cdot M^{r})/(\sum^{T'}\sum^{V}M^{r}),\\
	g^{r} &= (\sum^{T'}\sum^{V}f^{r}_{m} \cdot \bar{M^{r}})/(\sum^{T'}\sum^{V}\bar{M^{r}}),
	\end{aligned}
	\end{equation}
	where $M^{r} \in \mathbb{R}^{T' \times V}$ is a [0, 1] mask indicates the corresponding cropped skeleton fragments, while $\bar{M^{r}} = \textbf{I}-M^{r}$ indicates the corresponding remaining skeleton sequences. 
	$\textbf{I}$ is unit matrix with shape $\mathbb{R}^{T'\times{V}}$.
	Then, we extend the contrastive pairs with the \trimmed~view and the \truncated~view.
	Following the criteria of constructing contrastive pairs in MoCov2 \cite{chen2020improved},
	for the \trimmed~embedding $p^{r}$, it is positive with corresponding key embeddings $k_{p^{r}}^{+}$ while negative with $g^{r}$ and the embeddings stored in the memory bank $\textbf{Q}$. 
	Therefore, the contrastive loss for the \trimmed~view $p^{r}$ can be written as:
	\begin{equation}
	\mathcal{L}_{p^{r}} = -\log(\frac{\exp(p^{r} \cdot k_{p^{r}}^{+} / \tau)}{\exp(p^{r} \cdot k_{p^{r}}^{+} / \tau) + Z(p^{r}) + \exp{(g^{r} \cdot p^{r} / \tau)}}).
	\end{equation}
	
	Similarly, the \truncated~embedding $g^{r}$ is positive with corresponding key embeddings $k_{g^{r}}^{+}$, while negative with $p^{r}$ and the embeddings store in the memory bank $\textbf{Q}$. Therefore, the contrastive loss for the \truncated~view $g^{r}$ can be written as:
	\begin{equation}
	\mathcal{L}_{g^{r}} = -\log(\frac{\exp(g^{r} \cdot k_{g^{r}}^{+} / \tau)}{\exp(g^{r} \cdot k_{g^{r}}^{+} / \tau) + Z(g^{r}) + \exp{(g^{r} \cdot p^{r} / \tau)}}).
	\end{equation}
    
    Since the \trimmed~view and the \truncated~view share context information during forward inference, and the action information they contain is incomplete, they provide hard samples for contrastive learning.
    The expanded contrastive samples provide extensive hard positive pairs such as $p^{r}$ and $k_{p^{r}}^{+}$ (as well as $g^{r}$ and $k_{g}^{+}$) and hard negative pairs such as $p^{r}$ and $g^{r}$, thus helping the model to learn better global and local spatio-temporal representations and improving the performance in downstream tasks.
    
	Considering that the \trimmed~view and the \truncated~view are symmetrical,
	we use the average of $\mathcal{L}_{p^{r}}$ and $\mathcal{L}_{g^{r}}$ as the overall loss of the $r_{th}$ mix branch, which can formulated as:
	\begin{equation}
	\mathcal{L}^{r}_{mix} = \frac{\mathcal{L}_{p^{r}} + \mathcal{L}_{g^{r}}}{2}.
	\end{equation}
	
	Finally, the loss used to optimize the encoder can be formulated as:
	\begin{equation}
	\mathcal{L} = \mathcal{L}_{info} + \sum^{R}_{r=1}{\lambda\mathcal{L}^{r}_{mix}},
	\end{equation}
    where $\lambda$ is a hyperparameter to balance the easy contrastive pairs and the hard contrastive pairs.

	\section{Experiments}
	\subsection{Datasets}
	In order to evaluate the effectiveness of the proposed method, we conduct experiments on three widely used datasets for skeleton-based action recognition. 
	
	\noindent\textbf{NTU RGB+D \cite{shahroudy2016ntu}}, denoted as NTU-60, is the most widely used dataset for skeleton-based action recognition. 
	It contains 60 action classes and 56,578 action instances which are performed by 40 performers.
	We follow the recommended evaluation protocols cross-subject (Xsub) and cross-view (Xview) to evaluate our method.
	
	\noindent\textbf{NTU RGB+D 120} \cite{liu2019ntu}, denoted as NTU-120, is the expansion of NTU RGB+D dataset in the number of performer and action categories.
	The scale of this dataset is improved to 120 action classes and 113,945 action instances.
	Two recommended evaluation protocols, cross-subject (Xsub) and cross-set (Xset) are used to evaluate our method.
	
	\noindent\textbf{PKU-MMD} \cite{liu2017pku} contains almost 20,000 action instances and 51 action classes.
	It consists of two subsets, where part II is more challenging than part I due to more noise caused by view variation.
	We evaluate our method on cross-subject benchmark of both subsets.
	

	\subsection{Experiments Settings}\label{sec:expset}
	To perform a fair comparison, we use the same data pre-processing with SkeletonCLR \cite{li20213d} and AimCLR \cite{guo2021contrastive} except for that we resize the length of skeleton sequences to 64 frames, rather than 50 frames. 
	This allows our \aug~ to be implemented more efficiently, since the temporal downsample ratio $\mathcal{R}$ is 4 of ST-GCN backbone.
	Thus, the temporal size of the final output feature is 16.
    The	batch size for both pretraining and downstream tasks is set to 128 for by default, except in specific cases.
    
    \noindent\textbf{Data Augmentation}. For skeleton sequence, a spatial augmentation $Shear$ together with a temporal augmentation $Temporal~ Crop$ is adopted to generate the query and key views.
    $A$ and $A'$ use the same combination of augmentations but with different parameters  due to the randomness. 
    
	(1) $Shear$: The shear augmentation is a linear transformation on the spatial dimension. 
	The shape of 3D coordinates of body joints is slanted with a random angle.
	The transformation matrix is defined as:
	\begin{equation}
	T_{s} = \left[
	\begin{array}{ccc}
	1      & a_{12} &a_{13}\\
	a_{21} &    1   &a_{23}\\
	a_{31} & a_{32} &1
	\end{array}
	\right],
	\end{equation}
	where $a_{12}$, $a_{13}$, $a_{21}$, $a_{23}$, $a_{31}$, $a_{32}$ are shear factors that randomly sampled from a uniform distribution $a_{ij} \sim U(-\beta,\beta)$, where
	$\beta$ is the shear amplitude. 
	Follow SkeletonCLR and AimCLR, we set $\beta=0.5$.
	The skeleton sequence is multiplied by the transformation matrix $T_{s}$ on the channel dimension.
	
	(2) $Temporal~Crop$: For the temporal skeleton sequence, specifically, we symmetrically pad some frames to the sequence and then randomly crop it to the original length, which increases the diversity while maintaining the distinction of original samples.
	The padding length is defined as $T/\gamma$, where $\gamma$ is the padding ratio and here we set $\gamma=6$.
	
	\noindent\textbf{Self-Supervised Pretext Training}. The baseline method of our \Method~is SkeletonCLR which follows the MoCov2 framework \cite{chen2020improved}.
	Therefore, the hyperparameters queue size $Q$, temperature $\tau$ in MoCov2 are important.
	In SkeletonCLR, $K=32768$ and $\tau=0.07$, while we found that $K=32768$ and $\tau=0.2$ is a better choice under our experiments settings.
	In most cases, our reproduction performs better than the original results reported on \cite{li20213d}.
	Thus, we use the results of our reproduction as our baseline.
	For the backbone, we adopt ST-GCN \cite{yan2018spatial}, but reduce the number of channels in each layer to 1/4 of the original settings and the final feature dimension is set to 128.
	For the optimizer, we use SGD with momentum (0.9) and weight decay (0.0001). The model is trained for 300 epochs with a learning rate of 0.1.
	The model is trained for 300 epochs with a learning rate of 0.1.
	For fair comparison, we also use three streams of skeleton sequences, \emph{i.e.,} joint, motion, and bone denoted as \textbf{J}, \textbf{M}, and \textbf{B}, respectively.
	The ensemble results are obtained from the score-level fusion with equal weights.
	
	\subsection{Evaluation Protocol}
	We compare our method with other methods under a variety of evaluation protocols, including KNN evaluation protocol, linear evaluation protocol, finetune protocol, and semi-supervised evaluation protocol.
	
	\noindent\textbf{KNN Evaluation Protocol}. A k-nearest neighbor (KNN) classifier without trainable parameters is used on the features extracted from the trained encoder.
	For all reported KNN results, $K=20$, and the temperature is set to 0.1.
	The results reflect the quality of the features learned by the encoder.
	
	\noindent\textbf{Linear Evaluation Protocol}. This is the most commonly used protocol for classification downstream task.
	Specifically, we append a classification head (a fully connected layer together with a softmax layer) after the pretrained encoder and train the network with the encoder fixed.
	An SGD with an initial learning rate of 3.0 is used to train the network for 100 epochs. 
	
	\noindent\textbf{Finetune Protocol}. We append a linear classification head to the pretrained encoder.
	And then, we use an SGD optimizer with initial learning rate (0.1), weight decay (0.0001) to train the whole network for 110 epochs.
	The learning rate is decayed by 10 at the $50_{th}$, the $70_{th}$, and the $90_{th}$ epoch.
	We also use 10-epoch warmup to improve the stability of the training process.
	
	\noindent\textbf{Semi-Supervised Evaluation Protocol}. This protocol follows the same settings as finetune protocol except for the scale of training datasets.
	Only 1\% or 10\% randomly selected labeled data are used to finetune the whole network. 
	On PKU-MMD Part II benchmark with 1\% labeled data, the batch size is set to 52, due to the limited data.
	An SGD with an initial learning rate of 0.1 (decreases by 10 at $80_{th}$ epoch) is used to optimize the whole network for 100 epochs.
	We also use 20-epoch warmup to improve the stability of the training process.
	
	\subsection{Ablation Study}
	In this section, we conduct ablation studies on different datasets with linear evaluation protocol to verify the effectiveness of different components of our method.

	\begin{table*}
		\caption{Linear evaluation results compared with SkeletonCLR on NTU-60, PKU-MMD, and NTU-120 datasets. ``$\Delta$'' represents the gain compared to SkeletonCLR using the same stream data. \textbf{J}, \textbf{M} and \textbf{B} indicate joint stream, motion stream, and bone stream, respectively. ``3s'' means three streams fusion. ``\dag'' indicates that results reproduced on our settings.}
		\centering
		\small
		\tabcolsep1.7mm
		\begin{tabular}{l|c|cc|cc|cc|cc|cc|cc}
			\toprule
			\multirow{3}{*}{Method} & \multirow{3}{*}{Stream} & \multicolumn{4}{c|}{NTU-60(\%)} & \multicolumn{4}{c|}{PKU-MMD(\%)} & \multicolumn{4}{c}{NTU-120(\%)}                         \\ \cline{3-14}
			&      & \multicolumn{2}{c|}{Xsub}          & \multicolumn{2}{c|}{Xview}          & \multicolumn{2}{c|}{part I}       & \multicolumn{2}{c|}{part II}     & \multicolumn{2}{c|}{Xsub} & \multicolumn{2}{c}{Xset} \\
			&      & acc.            & $\Delta$         & acc.             & $\Delta$         & acc.            & $\Delta$      &acc.&$\Delta$& acc.            & $\Delta$       & acc.              & $\Delta$       \\ \midrule
			SkeletonCLR \cite{li20213d}  &\textbf{J}                  &68.3&&76.4&&80.9&&35.2&&56.8&&55.9& \\
			SkeletonCLR$^{\dag}$ \cite{li20213d}  &\textbf{J} & 74.8&  \textcolor{red}{+6.5}& 78.9 &  \textcolor{red}{+2.3}  & 81.1       & \textcolor{red}{+0.2}  &35.8 & \textcolor{red}{+0.6}  & 63.2            &       \textcolor{red}{+6.4}     & 58.9              & \textcolor{red}{+3.0} \\
			\textbf{\Method~(Ours)} &\textbf{J}& 79.6 & \textcolor{red}{+11.3}& 84.4& \textcolor{red}{+8.0}& \textbf{89.2}& \textcolor{red}{+8.3} & 51.6 &\textcolor{red}{+16.4} & 67.4   & \textcolor{red}{+10.6} &   \textbf{69.6}   & \textcolor{red}{+13.7} \\
			\textbf{\Methodplus~(Ours)} &\textbf{J}& \textbf{80.7} & \textcolor{red}{+12.4}& \textbf{85.5}& \textcolor{red}{+9.1}& 88.1& \textcolor{red}{+7.2} & \textbf{55.0} &\textcolor{red}{+19.8} & \textbf{69.0}   & \textcolor{red}{+12.2} &   68.2   & \textcolor{red}{+12.3} \\\midrule
			SkeletonCLR \cite{li20213d}    & \textbf{M}   & 53.3  &   & 50.8  &  & 63.4  &  &13.5 &  & 39.6 &  & 40.2 & \\
			SkeletonCLR$^{\dag}$ \cite{li20213d}  & \textbf{M}  & 49.6&   \textcolor{green}{-3.7}& 53.6&\textcolor{red}{+2.8}& 63.9& \textcolor{red}{+0.5}  & 16.8 & \textcolor{red}{+3.3}  & 41.3    &  \textcolor{red}{+1.7}              & 44.1              & \textcolor{red}{+3.9}\\
			\textbf{\Method~(Ours)}     & \textbf{M}&70.3& \textcolor{red}{+17.0}& \textbf{76.1} & \textcolor{red}{+25.3}& 81.5&\textcolor{red}{+18.1} & 32.1 &\textcolor{red}{+18.6} & \textbf{49.7}& \textcolor{red}{+10.1} &\textbf{53.8}   & \textcolor{red}{+13.6} \\ 
			\textbf{\Methodplus~(Ours)}     & \textbf{M}&\textbf{74.1}& \textcolor{red}{+20.8}& 74.8 & \textcolor{red}{+24.0}& \textbf{83.8}&\textcolor{red}{+20.4} & \textbf{32.4} &\textcolor{red}{+18.9} & 48.5& \textcolor{red}{+8.9} &50.5   & \textcolor{red}{+10.3} \\ \midrule
			SkeletonCLR \cite{li20213d}    & \textbf{B}     & 69.4  &  & 67.4  &  & 72.6  &  & 30.4 &  & 48.4 &  & 52.0 & \\
			SkeletonCLR$^{\dag}$ \cite{li20213d}    & \textbf{B}& 70.3 &   \textcolor{red}{+0.9} & 72.4 & \textcolor{red}{+5.0}  & 80.0 & \textcolor{red}{+7.4}  & 25.0 & \textcolor{green}{-5.4}   & 54.2            &    \textcolor{red}{+5.8}            & 58.7              & \textcolor{red}{+6.7}\\
			\textbf{\Method~(Ours)}     & \textbf{B}& 76.3& \textcolor{red}{+6.9}& 82.0& \textcolor{red}{+14.6}& 89.0 & \textcolor{red}{+16.4} & 41.8 &\textcolor{red}{+11.4} & \textbf{67.1}& \textcolor{red}{+18.7}&\textbf{63.1}& \textcolor{red}{+11.1} \\
			\textbf{\Methodplus~(Ours)}     & \textbf{B}& \textbf{79.1}& \textcolor{red}{+9.7}& \textbf{82.6}& \textcolor{red}{+15.2}& \textbf{89.1} & \textcolor{red}{+16.5} & \textbf{46.0} &\textcolor{red}{+15.6} & 63.0& \textcolor{red}{+14.6}&60.7& \textcolor{red}{+8.7} \\ \midrule
			3s-SkeletonCLR \cite{li20213d} & \textbf{J}+\textbf{M}+\textbf{B}                 &75.0&&79.8&&85.3&&40.4&&60.7&&62.6& \\
			3s-SkeletonCLR$^{\dag}$ \cite{li20213d}   &\textbf{J}+\textbf{M}+\textbf{B}      &75.9&\textcolor{red}{+0.9}&79.8&\textcolor{green}{0.0}&85.4&\textcolor{red}{+0.1}&37.6& \textcolor{green}{-2.8}&65.0&\textcolor{red}{+4.3}&65.9& \textcolor{red}{+3.3}\\
			\textbf{3s-\Method~(Ours)}  & \textbf{J}+\textbf{M}+\textbf{B} &81.0&\textcolor{red}{+6.0}&85.6&\textcolor{red}{+5.8} & 90.6&\textcolor{red}{+5.3}&52.9&\textcolor{red}{+12.5}&69.1&\textcolor{red}{+8.4}&69.9&\textcolor{red}{+7.3}\\
			\textbf{3s-\Methodplus~(Ours)}  & \textbf{J}+\textbf{M}+\textbf{B} &\textbf{82.7}&\textcolor{red}{+7.7}&\textbf{87.1}&\textcolor{red}{+7.3} & \textbf{91.1}&\textcolor{red}{+5.8}&\textbf{57.1}&\textcolor{red}{+16.7}&\textbf{70.5}&\textcolor{red}{+9.8}&\textbf{70.7}&\textcolor{red}{+8.1}\\ \bottomrule
		\end{tabular}
		\label{tab:com}
	\end{table*}
	
	\noindent\textbf{Comparisons with SkeletonCLR}. 
	We conduct experiments on NTU-60, NTU-120, and PKU-MMD datasets to compare our method with baseline method SkeletonCLR in detail.
	As can be seen from Table \ref{tab:com}, the reproduced \Method~with our settings achieves better performance than the original one, so we use the reproduced one as our baseline for fair comparison.
	Moreover, our \Method~substantially improves SkeletonCLR, especially for the motion stream and bone stream.
	Multiple \aug~augmentation strategy could further boost the performance, and 3s-\Methodplus~achieves the best performance compared with SkeletonCLR and 3s-\Method.
	The experimental results demonstrate the effectiveness of our proposed method.
	
	\noindent\textbf{Choice of Cropped Skeleton Fragments}.
	There are four hyperparameters ($\mathcal{B}_{sl} {\in} (0, 5)$, $\mathcal{B}_{su} {\in} (0, 5)$, $\mathcal{B}_{tl} {\in} (0,16)$, and $\mathcal{B}_{tu} {\in} (0, 16)$) introduced by our method.
	To determine them, we first fix $\mathcal{B}_{tl}{=}5$ and $\mathcal{B}_{tu}{=}11$, and try all the valid combinations between $\mathcal{B}_{sl}$ and $\mathcal{B}_{su}$.
	As show in Table \ref{tab:bodundries}, we found that the best setting is $\mathcal{B}_{sl}{=}2$ and $\mathcal{B}_{su}{=}3$.
	Then we fix the best combination of $\mathcal{B}_{sl}{=}2$, $\mathcal{B}_{su}{=}3$, and $\mathcal{B}_{tu}{=}11$ to search for the best $\mathcal{B}_{tl}$.
	Finally, we search for the best $\mathcal{B}_{tu}$ with other three parameters fixed. 
	Based on the results shown in Table \ref{tab:bodundries}, we choose $\mathcal{B}_{sl}{=}2$, $\mathcal{B}_{su}{=}3$, $\mathcal{B}_{tl}{=}7$, and $\mathcal{B}_{tu}{=}11$ as our default setting to perform \aug~augmentation.

	\begin{table}[t]
		\caption{Boundaries search experiments on joint stream of NTU-60 dataset.}
		\centering
		\small
		\begin{tabular}{cccc|ccc}
			\toprule
			&                    &                     &             & \multicolumn{3}{c}{NTU-60-J (\%)} \\
			$\mathcal{B}_{sl}$ & $\mathcal{B}_{su}$ & $\mathcal{B}_{tl}$  & $\mathcal{B}_{tu}$   & Xsub      & Xview  & Avg\\ \midrule
			-         & -        & -     & -       & 74.8           & 78.9 & 76.9\\
			1         & 1        & 5     & 11      & 78.2           & 80.6 & 79.4\\
			1         & 2        & 5     & 11      & 79.2           & 79.8 & 79.5\\
			1         & 3        & 5     & 11      & 78.1           & 81.7 & 79.9\\ 
			1         & 4        & 5     & 11      & 77.8           & 79.7 & 78.8\\
			2         & 2        & 5     & 11      & 77.8           & 83.1 & 80.5\\
			2         & 3        & 5     & 11      & 78.8           & 84.2 & \textbf{81.5}\\
			2         & 4        & 5     & 11      & 78.0           & 83.9 & 81.0\\
			3         & 3        & 5     & 11      & 76.6           & 83.8 & 80.2\\
			3         & 4        & 5     & 11      & 78.3           & 83.2 & 80.8\\
			4         & 4        & 5     & 11      & 77.2           & 82.4 & 79.8\\  \midrule
			2         & 3        & 1     & 11      & 77.8           & 82.4 & 80.1\\
			2         & 3        & 3     & 11      & 80.4           & 82.0 & 81.2\\
			2         & 3        & 5     & 11      & 78.8           & 84.2 & 81.5\\
			2         & 3        & 7     & 11      & 79.6           & 84.4 & \textbf{82.0}\\ 
			2         & 3        & 9     & 11      & 79.0           & 83.1 & 81.1\\ \midrule
			2         & 3        & 7     & 7       & 79.9           & 82.9 & 81.4\\
			2         & 3        & 7     & 9       & 79.8           & 84.0 & 81.9\\ 
			2         & 3        & 7     & 11      & 79.6           & 84.4 & \textbf{82.0}\\
			2         & 3        & 7     & 13      & 79.3           & 83.4 & 81.4\\
			2         & 3        & 7     & 15      & 79.3           & 83.0 & 81.2\\ \bottomrule
		\end{tabular}
		\label{tab:bodundries}
	\end{table}
	
	\noindent\textbf{Balance Easy Contrastive Pairs and Hard Contrastive Pairs.}
	In our model, we present that the power of easy contrastive pairs and hard contrastive pairs are traded by a hyperparameter $\lambda$.
	Here, we analyze how $\lambda$ affects the performance of the model.
	For convenience, we set $R{=}1$, and compare the performance of different $\lambda$ on NTU-60 dataset with linear evaluation protocol.
	From Table \ref{tab:lambda}, when $\lambda{=}1.0$, the model get the highest recognition accuracy, showing large improvements than cases with $\lambda{=}0.1$ and $\lambda{=}10.0$.
	Therefore, we set $\lambda{=}1.0$ as default setting.
	
	\begin{table}[t]
		\caption{Action recognition accuracies with different $\lambda$}
		\centering
		\small
		\begin{tabular}{c|ccc}
			\toprule
			      & \multicolumn{3}{c}{NTU-60-J (\%)} \\
			$\lambda$        & Xsub      & Xview  & Avg\\ \midrule
			0.1 & 75.1& 81.3& 78.2 \\
			1.0 & 79.6& 84.4 & \textbf{82.0}\\
			10.0 & 83.0& 76.8& 79.9\\
			\bottomrule
		\end{tabular}
		\label{tab:lambda}
	\end{table}
		
	\begin{table}[t]
		\caption{Linear evaluation results on NTU-60 dataset for different training epochs. ``\dag'' indicates that results are reproduced on our settings.}
		\centering
		\small
		\begin{tabular}{c|c|ccc}
			\toprule
			\multicolumn{2}{c|}{Method} & 100ep & 200ep & 300ep \\ \midrule
			\multirow{2}{*}{Xview} & SkeletonCLR$^{\dag}$ \cite{li20213d} & 75.5 & 77.8 & 78.9  \\
			& \Method~(Ours)                   & 81.3 & 83.5 & 84.4  \\ \midrule
			\multirow{2}{*}{Xsub} & SkeletonCLR$^{\dag}$ \cite{li20213d}  & 71.1 & 73.9 & 74.8  \\
			& \Method~(Ours)                   & 78.2 & 79.3 & 79.6  \\ \bottomrule					
		\end{tabular}
		\label{tab:epochs}
	\end{table}
	
	\noindent\textbf{Training with Different Epochs}. 
	We also investigate the influence of different training epochs. 
	The results are shown in Table~\ref{tab:epochs}, as we can see that all methods are close to convergence, thus we believe that 300 epochs are sufficient for comparing.
	Our method outperforms the baseline method SkeletonCLR with all settings of different training epochs.
	It is worth mentioning that with only 100 training epochs, the proposed \Method~outperforms SkeletonCLR with 300 training epochs.
	The results not only show good property of convergence brought by \Method, but also verify that our method can enhance the representation capacity of the model with more training epochs.

	\noindent\textbf{Effectiveness of Proposed \aug~Strategy}.
	Our \aug~augmentation strategy utilizes the topological information to perform skeleton sequence mixing operation, which not only makes better use of the augmented data, but also maintains the consistency of action information, thus providing more reasonable and informative features for contrastive learning.
	To validate the efficiency of our method, we compare \aug~with zeros padding strategy, which pads zeros to the remaining skeleton sequences.
	The comparisons are shown in Table \ref{tab:padding}.
	Zeros padding strategy slightly improves the performance, while a big improvement by $4.8\% {\sim} 5.5\%$ has been achieved with our \aug.
	The results demonstrate that our \aug~augmentation can make full use of the augmented skeleton sequences to provide more informative features for contrastive learning.
	Moreover, different from images, skeleton data contains topological information and our method utilizes such information to perform part-level mixing operation which maintains the consistency of both remaining skeleton sequences and cropped skeleton fragments.
	To verify the efficiency of the topological information, we compare our method with the random strategy, which randomly selects some skeleton joints to crop.
	From Table~\ref{tab:spatial}, our method outperforms the random strategy especially on NTU-60 Xivew benchmark, which demonstrates the effectiveness of the topological information and our method can make good use of such information to help the model learn better spatio-temporal representations.

	\begin{table}[t]
		\centering
		\parbox{.4\linewidth}{
		\caption{Comparison with zeros padding strategies.}
			\centering
			\footnotesize
			\setlength{\tabcolsep}{6pt}
			\begin{tabular}{c|cc}
				\hline
				NTU-60-J & Xsub          & Xview          \\ \hline
				Baseline   & 74.8          & 78.9           \\
				Zeros      & 76.2          & 81.1           \\
				\aug     & \textbf{79.6} & \textbf{84.4}  \\ \hline
			\end{tabular}
			\label{tab:padding}
		}\hfill
		\parbox{.55\linewidth}{
			\caption{Comparison between different cropping strategies.}
			\centering
			\footnotesize
			\setlength{\tabcolsep}{6pt}
			\begin{tabular}{c|cc}
				\hline
				NTU-60-J & Xsub          & Xview          \\ \hline
				Baseline   & 74.8          & 78.9           \\
				Random     & 77.7          & 80.2           \\
				Topology   & \textbf{79.6} & \textbf{84.4}  \\ \hline
			\end{tabular}
			\label{tab:spatial}
		}\hfill
	\end{table}	
		
	\begin{table}[t]
		\caption{Ablation study results on NTU-60 dataset.}
		\centering
		\small
		\tabcolsep1.5mm
		\begin{tabular}{ccccc|cc}
			\toprule
			$Cont$ & $\mathcal{L}_{p}$ & $\mathcal{L}_{g}$ & $d_{pg}$ & $detach$ &   Xsub              & Xview           \\ \midrule
			\xmark &\xmark         & \xmark      & \xmark    & \xmark   & 74.8             & 78.9            \\
			\cmark & & & & &75.2 & 79.4           \\
			&\cmark          &       &    &    & 78.4             & 79.8            \\
			 &         & \cmark      &    &    & 77.4             & 82.3            \\
			&\cmark          & \cmark      &    &    & 79.1             & 82.4            \\
			&\cmark          & \cmark      & \cmark   &    & 79.3             & 84.3             \\
			&\cmark          & \cmark      & \cmark   & \cmark   & \textbf{79.6}    & \textbf{84.4}            \\ \bottomrule
		\end{tabular}
		\label{tab:ablation}
	\end{table}
	
	\noindent\textbf{Effectiveness of Hard Contrastive Pairs}.
	\begin{figure}[t]
		\centering
		\includegraphics[width=1\linewidth]{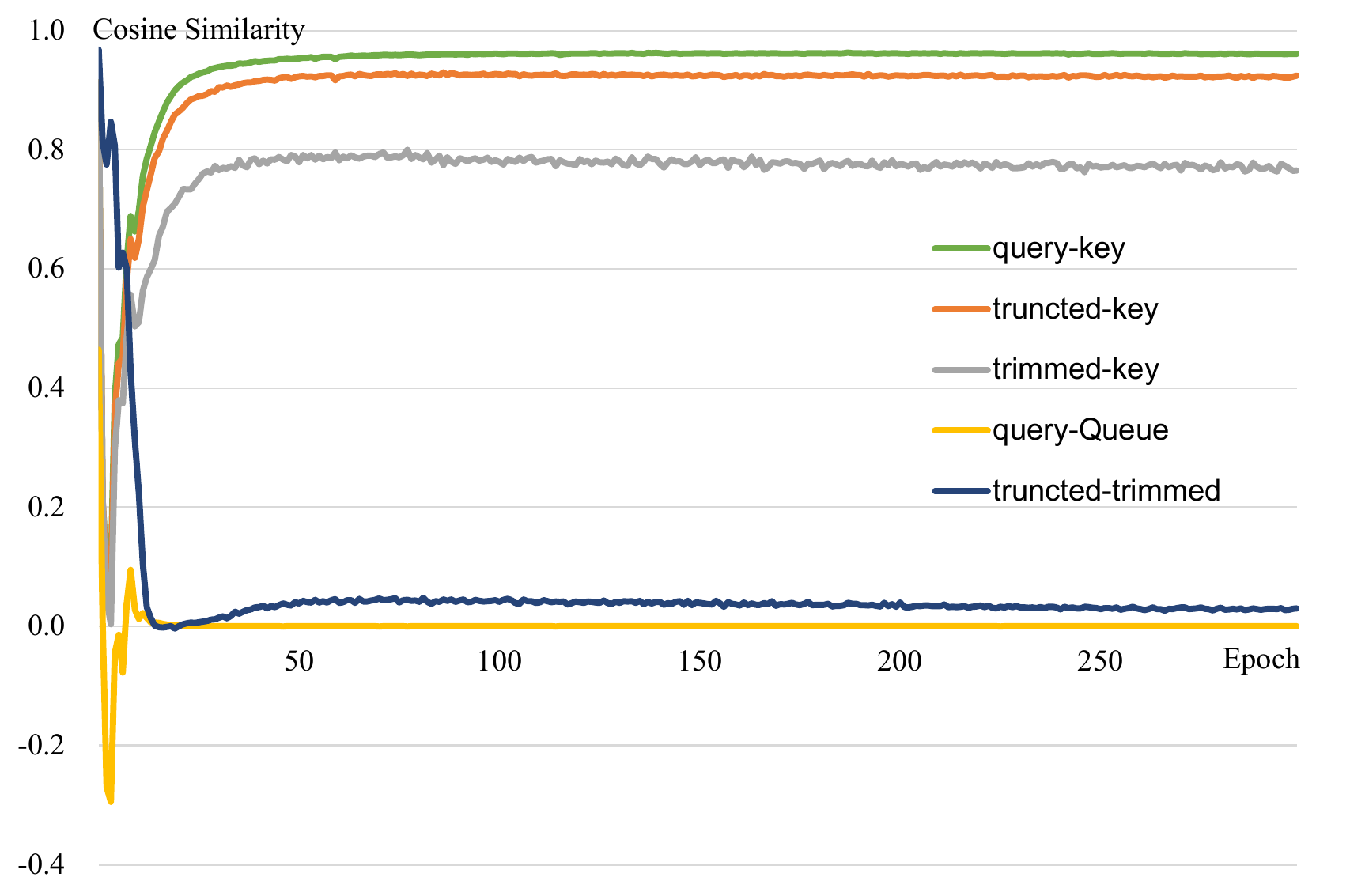} 
		\caption{Comparison of average similarities over all samples between different views. The model is trained on NTU-60 Xview benchmark.}
		\label{fig:sim}
	\end{figure}
	Our method can provide abundant hard contrastive pairs in contrastive pretext task.
	To further verify the effectiveness of our method, we test the influence of each component.
	A blank control method (denoted as $Cont$) is constructed by naively extending a query view and constructing the corresponding contrastive pairs.
	The results are shown in Table \ref{tab:ablation}, where $\mathcal{L}_{p}$ and $\mathcal{L}_{g}$ denote whether to use $p$ and $g$ to construct hard contrastive pairs, respectively.
	$d_{pg}$ denotes whether to use $p$ and $g$ to construct hard negative pairs.
	Since SimSiam \cite{chen2021exploring} further finds that stop-gradient is critical to prevent from collapsing, we also use this strategy denoted as $detach$ when constructing hard negative pairs between $p$ and $g$.
	As shown in Table \ref{tab:ablation}, naively extending contrastive pairs based on normal augmentations brings limited returns in terms of performance (0.4\% on Xsub benchmark and 0.5\% on Xview benchmark), which indicates that more does not mean effective.
	Compared with the blank control method, we found that the hard positive pairs contribute a lot to the performance and using hard negative pairs also greatly improves the performance on NTU-60 Xview benchmark.
	The stop-gradient strategy can further improve the performance.

	To further analyze our approach, we compare the average cosine similarity of easy contrastive pairs provided by baseline and hard contrastive pairs introduced by our method.
	As shown in Figure~\ref{fig:sim}, as the training progresses, the cosine similarity between easy positive contrastive pairs (query and key) is very close to 1, and the cosine similarity between easy positive contrastive pairs (query and Queue) is very close to 0, which contribute less and less to the loss \cite{kalantidis2020hard}.
	Hard positive contrastive pairs (\truncated~and key, \trimmed~and key) and hard negative contrastive pairs (\truncated~and \trimmed) are harder than easy contrastive pairs, which contributes more to optimize the model.
	
	To summarize the above, our method uses \aug~augmentation strategy to generate the mixed skeleton sequences and separate the \trimmed~view and the \truncated~view at the feature level to provide hard samples for contrastive learning.
	\Method~further utilizes the hard samples to expand the contrastive pairs with hard positive pairs and hard negative pairs, which significantly improves the capacity of the model in learning better complete and discriminative spatio-temporal representations for skeleton-based action recognition.

    \begin{table}[t]
		\caption{Action recognition accuracies with different $R$}
		\centering
		\small
		\begin{tabular}{c|ccc}
			\toprule
			      & \multicolumn{3}{c}{NTU-60-J (\%)} \\
			$R$        & Xsub      & Xview  & Avg\\ \midrule
			1 & 79.6 & 84.4 & 82.0\\
			2 & 80.9 & 85.0 & 83.0 \\
			3 & 80.7 & 85.5 & \textbf{83.1}\\
			4 & 80.5 & 79.3 & 79.9\\
			\bottomrule
		\end{tabular}
		\label{tab:R}
	\end{table}
	\begin{table}[t]
		\caption{Comparison with SkeletonCLR and AimCLR under KNN evaluation protocol with joint stream. ``\dag'' indicates that results reproduced on our settings.}
		\centering
		\small
		\tabcolsep0.9mm
		\begin{tabular}{l|cccccc}
			\toprule
			\multirow{2}{*}{Method} & \multicolumn{2}{c}{NTU-60} & \multicolumn{2}{c}{NTU-120} &  \multicolumn{2}{c}{PKU-MMD}\\
			& Xsub           & Xview           & Xsub            & Xset           & part I & part II \\ \midrule
			SkeletonCLR$^{\dag}$ \cite{li20213d}    & 64.8           & 60.7            & 41.9            & 42.9   & 64.9 & 19.9         \\
			AimCLR$^{\dag}$ \cite{guo2021contrastive} & 71.0 & 63.7 & 48.9 & 47.3 & 73.2 & 19.4 \\
			\textbf{\Method~(Ours)}      & \textbf{72.3}  & \textbf{65.5}      & \textbf{49.3}   & \textbf{48.3}  & \textbf{75.7} & \textbf{33.8} \\ \bottomrule
		\end{tabular}
		\label{tab:KNN}
	\end{table}
	\begin{table}[h!]
		\caption{Linear evaluation results on NTU-120 dataset.}
		\centering
		\small
		\begin{tabular}{l|cc}
			\toprule
			Method                            & Xsub (\%)       & Xset (\%)        \\      \midrule
			\emph{Single-stream:}             &                &                 \\
			PCRP \cite{xu2021prototypical}                 & 41.7          & 45.1           \\
			AS-CAL \cite{rao2021augmented}  & 48.6           & 49.2            \\
			SkeletonCLR (\cite{li20213d})      & 56.8 & 55.9         \\
			ISC \cite{thoker2021skeleton}                 & 67.9  & 67.1            \\ 
			AimCLR (\cite{guo2021contrastive}) & 63.4 & 63.4 \\
			\textbf{\Method~(Ours)}      & 67.4  & \textbf{69.6}       \\
			\textbf{\Methodplus~(Ours)}      & \textbf{69.0}  & 68.2      \\ \hline
		    \emph{Multi-stream:} & & \\
		    3s-SkeletonCLR (\cite{li20213d}) & 60.7 & 62.6 \\
			3s-CrosSCLR (\cite{li20213d})      & 67.9           & 66.7            \\
			3s-AimCLR \cite{guo2021contrastive}             & 68.2           & 68.8            \\
			\textbf{3s-\Methodplus~(Ours)}   & 69.1  & 69.9    \\
			\textbf{3s-\Methodplus~(Ours)}      & \textbf{70.5}  & \textbf{70.7}       \\ \bottomrule
		\end{tabular}
		\label{tab:ntu120-le}
	\end{table}			

    \noindent\textbf{Influence of The Multiple \aug~Augmentation Strategy}
    As described in Sec.~\ref{sec:skelemix} and Sec.~\ref{sec:skelemixclr}, to provide a more adequate hard contrastive pairs, we propose to perform multiple \aug~augmentations.
    Here, we conduct experiments on NTU-60 datasets with linear evaluation protocol to analyze how $R$ affects the performance.
    As shown in Table~\ref{tab:R}, when $R{=}2$, the average performance over NTU-60 Xsub and Xview benchmarks is improved from 82.0 to 83.0.
    There is a slight improvement in average performance when increasing $R$ from 2 to 3.
    While increasing $R$ from 3 to 4 leads to a dramatic drop.
    We believe this is caused by the model focusing too much on hard contrastive pairs and destroying the balance between simple and hard contrastive pairs.
    Based on the above experiments, we set $R{=}3$ as the default setting.

	\subsection{Comparison with State-of-the-Art Methods}
	
	\noindent\textbf{KNN Evaluation Protocol Results.} 
	We compare our \Method~with SkeletonCLR and AimCLR on three datasets.
	The results are shown in Table \ref{tab:KNN}, our \Method~outperforms both methods by a large margin especially in relatively small dataset PKU-MMD, which shows that the features learned by \Method~are more discriminative.
	
	\noindent\textbf{Linear Evaluation Protocol Results.} 
	Table \ref{tab:ntu120-le},  \ref{tab:ntu-le}, and \ref{tab:pku-le} show the comparisons on NTU-120, NTU-60, and PKU-MMD datasets with linear evaluation protocol, respectively.
	Notably, our single \Method~outperforms some ensemble methods such as 3s-SkeletonCLR \cite{li20213d}, 3s-Colorization \cite{yang2021skeleton}, 3s-CrosSCLR \cite{li20213d}, and 3s-AimCLR \cite{guo2021contrastive} on NTU-60, PKU-MMD, and NTU-120 Xset benchmarks.
	Multiple \aug~augmentation strategy could further boost the performance, achieving considerable gains on most benchmarks.
	It is worth mentioning that just with only single joint stream, our \Method or \Methodplus surpasses many multi-stream methods with a big margin.
	When multiple streams of information are introduced, our method can be further improved. 
	On NTU-60, our 3s-\Methodplus~outperforms 3s-AimCLR by 3.8\% on Xsub and outperforms ISC \cite{thoker2021skeleton} by 1.9\% on Xview.
	On NTU-120, our 3s-\Methodplus~outperforms 3s-AimCLR by 2.3\% and 1.9\% on Xsub and Xset, respectively.
	On PKU-MMD, our 3s-\Methodplus~outperforms 3s-AimCLR by a large margin (3.3\% on part I and 18.6\% on part II).
	Based on the above, our method performs well on both large scale and small scale datasets, which demonstrates the effectiveness and generalization of our method.
	
	\begin{table}[t]
		\caption{Linear evaluation results on NTU-60 dataset.}
		\centering
		\small
		\begin{tabular}{l|cc}
			\toprule
			Method  & Xsub (\%)          & Xview (\%)           \\ \midrule
			\emph{Single-stream:}             &                &                 \\
			LongT GAN \cite{zheng2018unsupervised}             & 39.1           & 48.1            \\
			MS$^2$L \cite{lin2020ms2l}          & 52.6           & -               \\
			PCRP \cite{xu2021prototypical}                 & 53.9          & 63.5           \\
			AS-CAL \cite{rao2021augmented} & 58.5           & 64.8            \\
			SkeletonCLR (\cite{li20213d})           & 68.3           & 76.4            \\
			ISC \cite{thoker2021skeleton}                 & 76.3           & 85.2      \\
			AimCLR \cite{guo2021contrastive}                & 74.3           & 79.7            \\ 
			\textbf{\Method~(Ours)}          & 79.6  & 84.4   \\
			\textbf{\Methodplus~(Ours)}          & \textbf{80.7}  & \textbf{85.5}   \\ \midrule
					\emph{Multi-stream:}              &                &                 \\
					3s-SkeletonCLR (\cite{li20213d})        & 75.0           & 79.8            \\
			3s-Colorization \cite{yang2021skeleton}       & 75.2           & 83.1            \\
			3s-CrosSCLR (\cite{li20213d})           & 77.8           & 83.4            \\
			3s-AimCLR \cite{guo2021contrastive}              & 78.9           & 83.8            \\ 
			\textbf{3s-\Method~(Ours)}       & 81.1  & 85.6   \\ 
			\textbf{3s-\Methodplus~(Ours)}       & \textbf{82.7}  & \textbf{87.1}   \\ \bottomrule
		\end{tabular}
		\label{tab:ntu-le}
	\end{table}
	
	\begin{table}[t]
		\caption{Linear evaluation results on PKU-MMD dataset.}
		\centering
		\small
		\begin{tabular}{l|cc}
			\toprule
			Method & part \uppercase\expandafter{\romannumeral1} (\%) & part \uppercase\expandafter{\romannumeral2} (\%) \\ \midrule
			\emph{Single-stream:}       &                &                 \\
			LongT GAN \cite{zheng2018unsupervised}         & 67.7           & 26.0            \\
			MS$^2$L \cite{lin2020ms2l}         & 64.9           & 27.6            \\
			SkeletonCLR \cite{li20213d} & 80.9 & 35.2 \\
			AimCLR \cite{guo2021contrastive} & 83.4 & 36.8 \\
			ISC \cite{thoker2021skeleton}             & 80.9           & 36.0            \\
			\textbf{\Method~(Ours)}      & \textbf{89.2}  & 51.6   \\
			\textbf{\Methodplus~(Ours)}      & 88.1  & \textbf{55.0}   \\
			\midrule
			\emph{Multi-stream:}       &                &                 \\
			3s-SkeletonCLR \cite{li20213d} & 85.3 & 40.4 \\
			3s-CrosSCLR (\cite{li20213d}) & 84.9           & 21.2            \\
			3s-AimCLR \cite{guo2021contrastive}          & 87.8           & 38.5            \\  
			\textbf{3s-\Method~(Ours)}      & 90.6  & 52.9   \\
			\textbf{3s-\Methodplus~(Ours)}  & \textbf{91.1}  & \textbf{57.1}   \\ \bottomrule
		\end{tabular}
		\label{tab:pku-le}
	\end{table}
			
	\begin{figure*}[t]
		\centering
		\subfloat[\Method-\textbf{J}]{\includegraphics[width=0.3\linewidth]{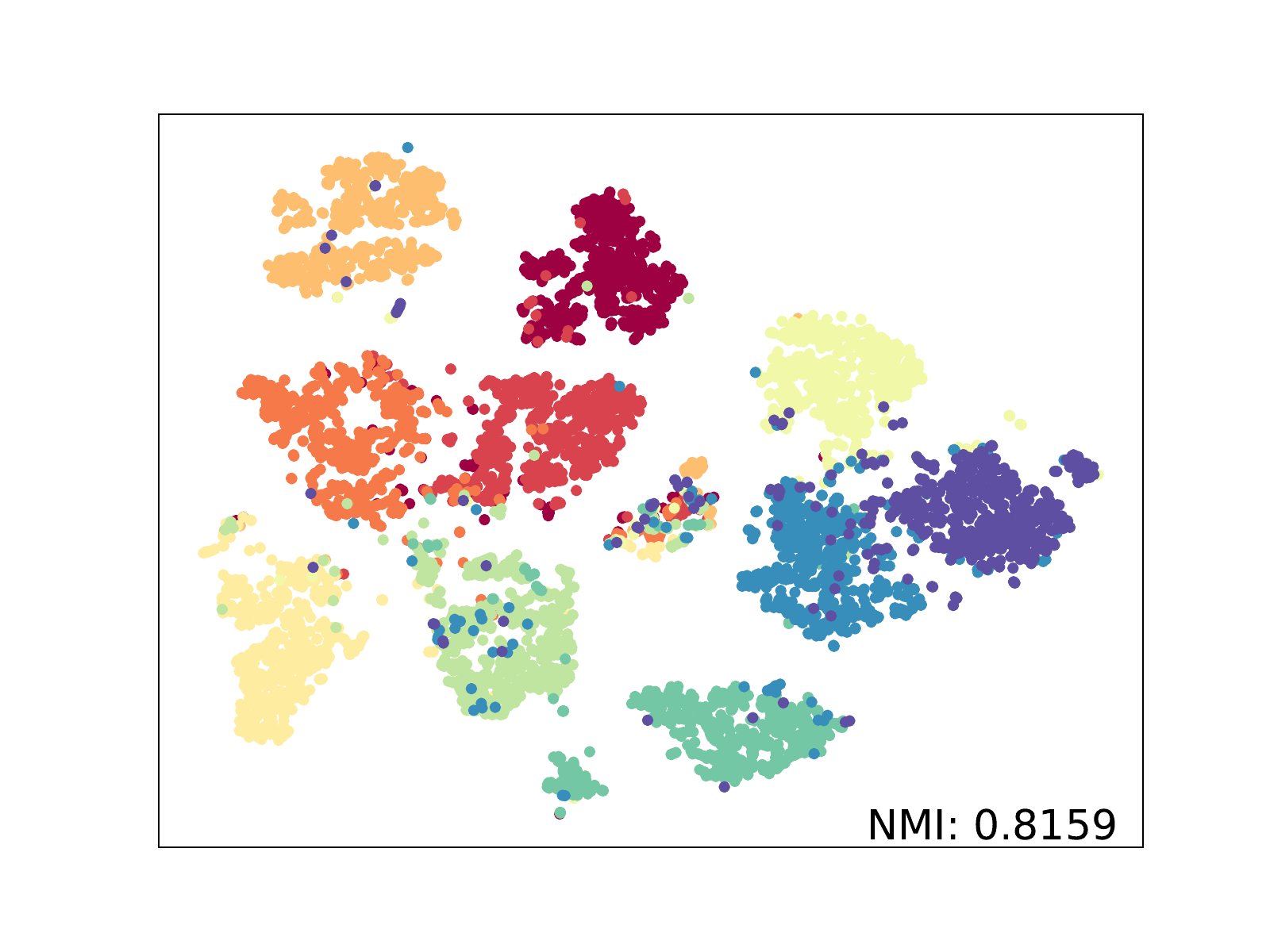}}
		\subfloat[\Method-\textbf{M}]{\includegraphics[width=0.3\linewidth]{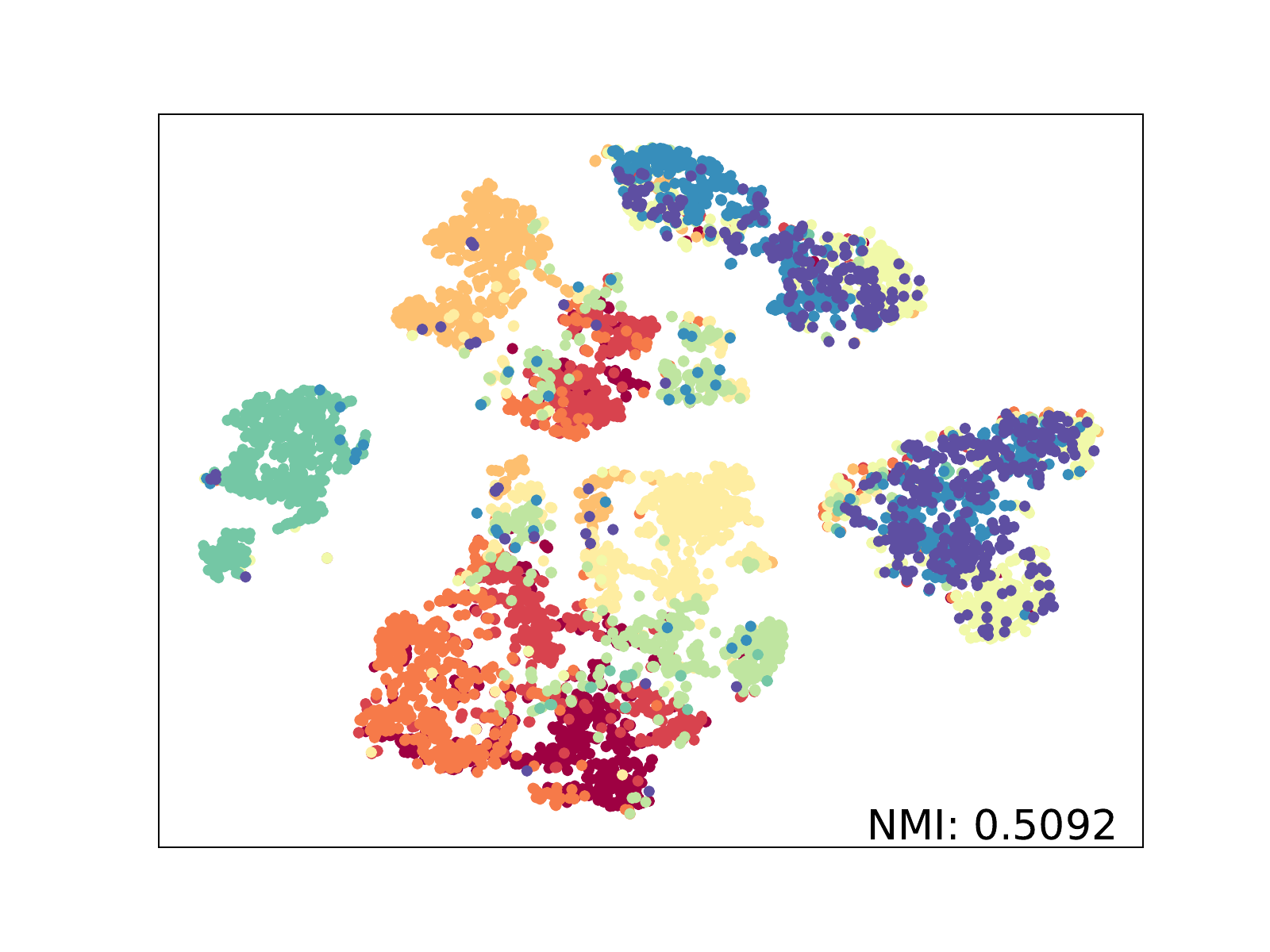}}
		\subfloat[\Method-\textbf{B}]{\includegraphics[width=0.3\linewidth]{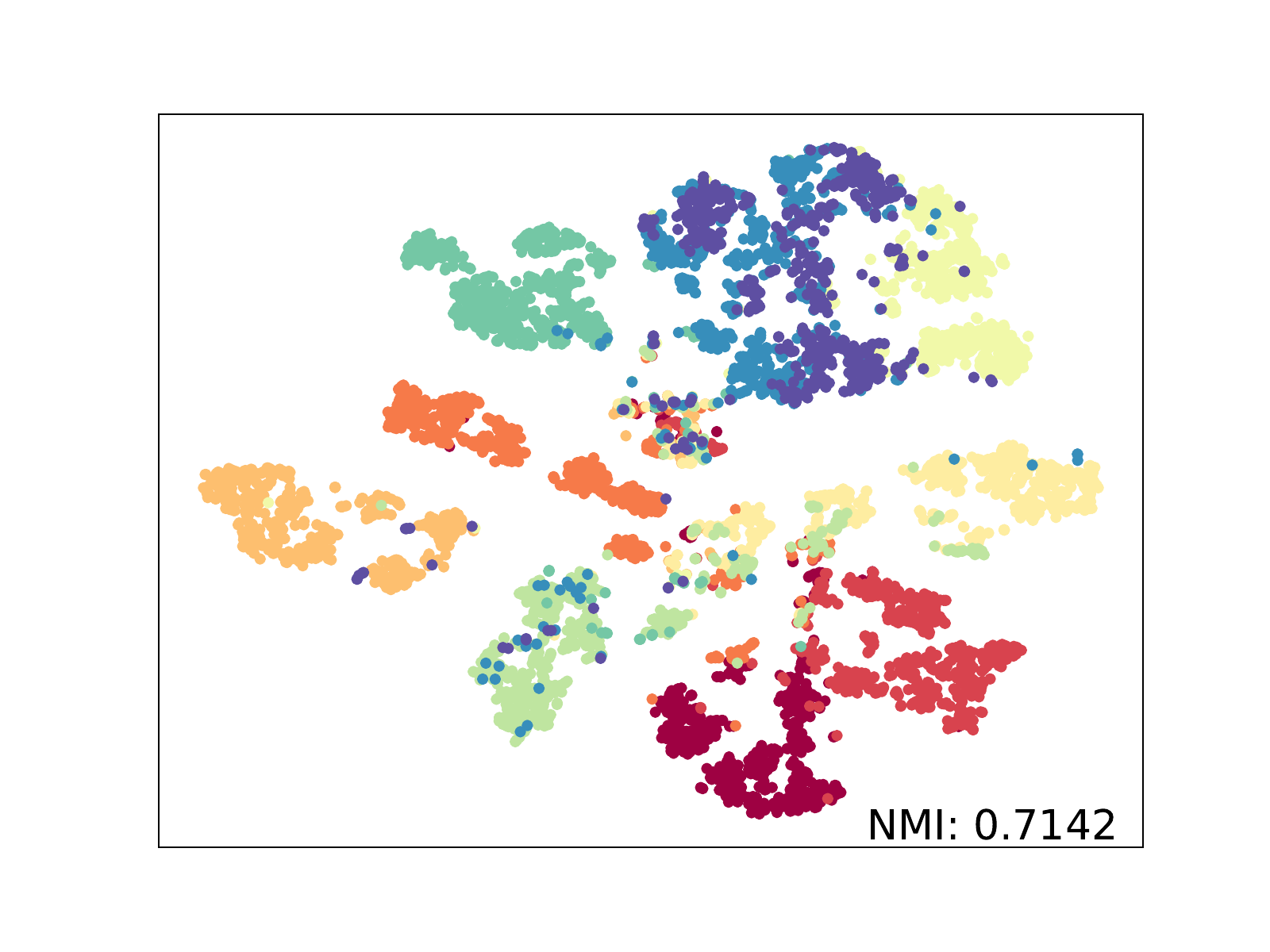}}
		\\
		\subfloat[SkeletonCLR-\textbf{J}]{\includegraphics[width=0.3\linewidth]{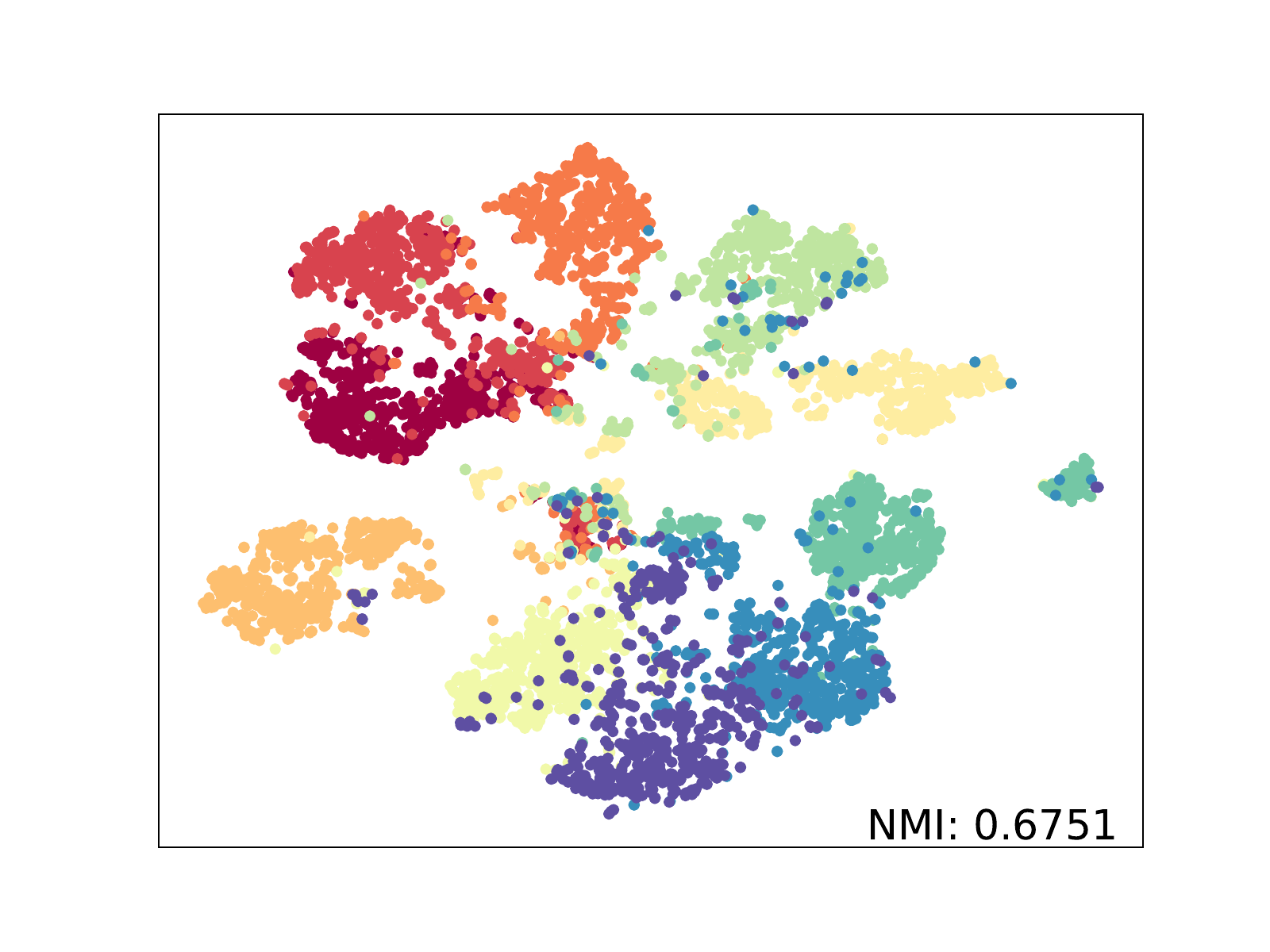}}
		\subfloat[SkeletonCLR-\textbf{M}]{\includegraphics[width=0.3\linewidth]{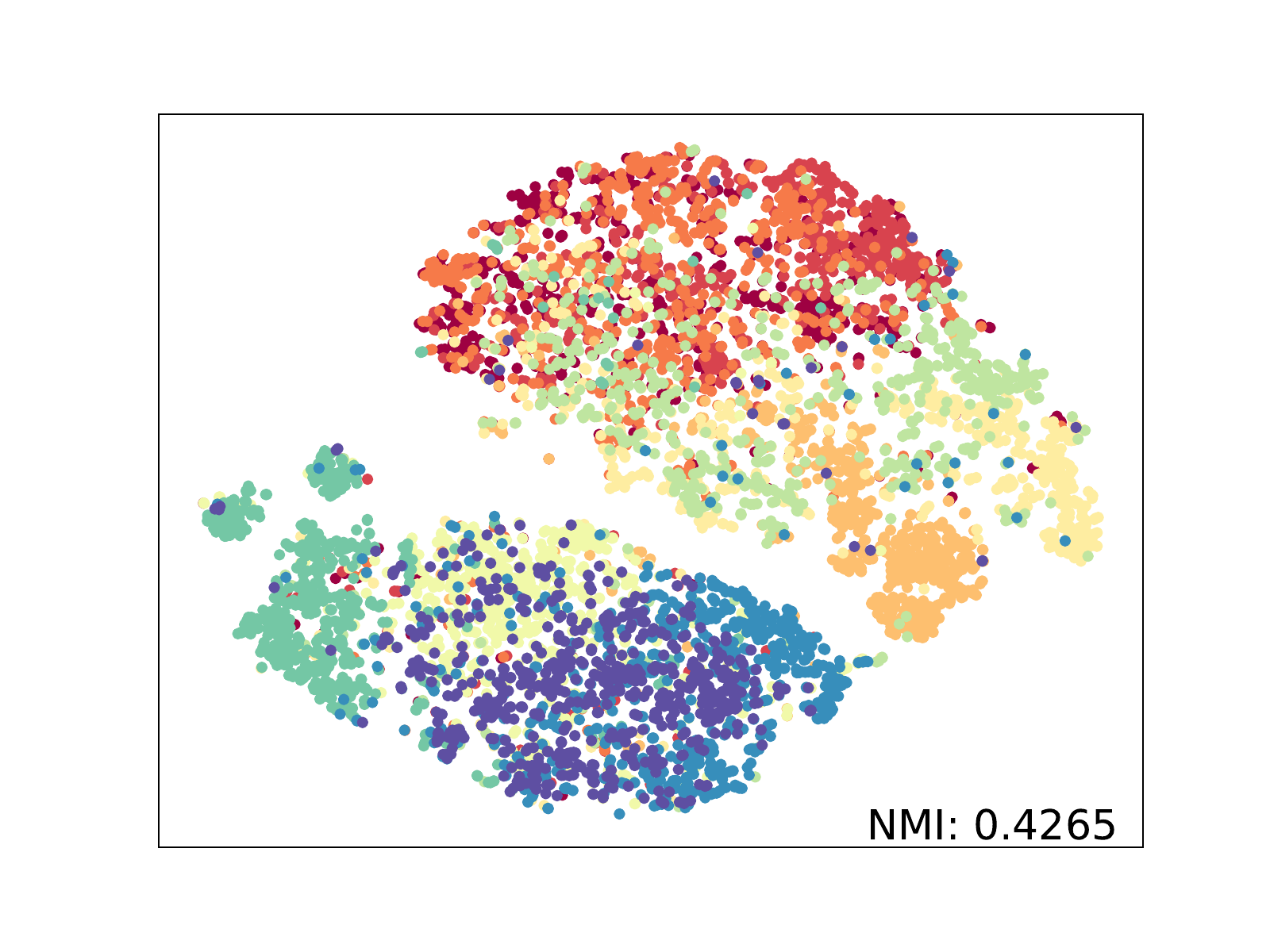}}
		\subfloat[SkeletonCLR-\textbf{B}]{\includegraphics[width=0.3\linewidth]{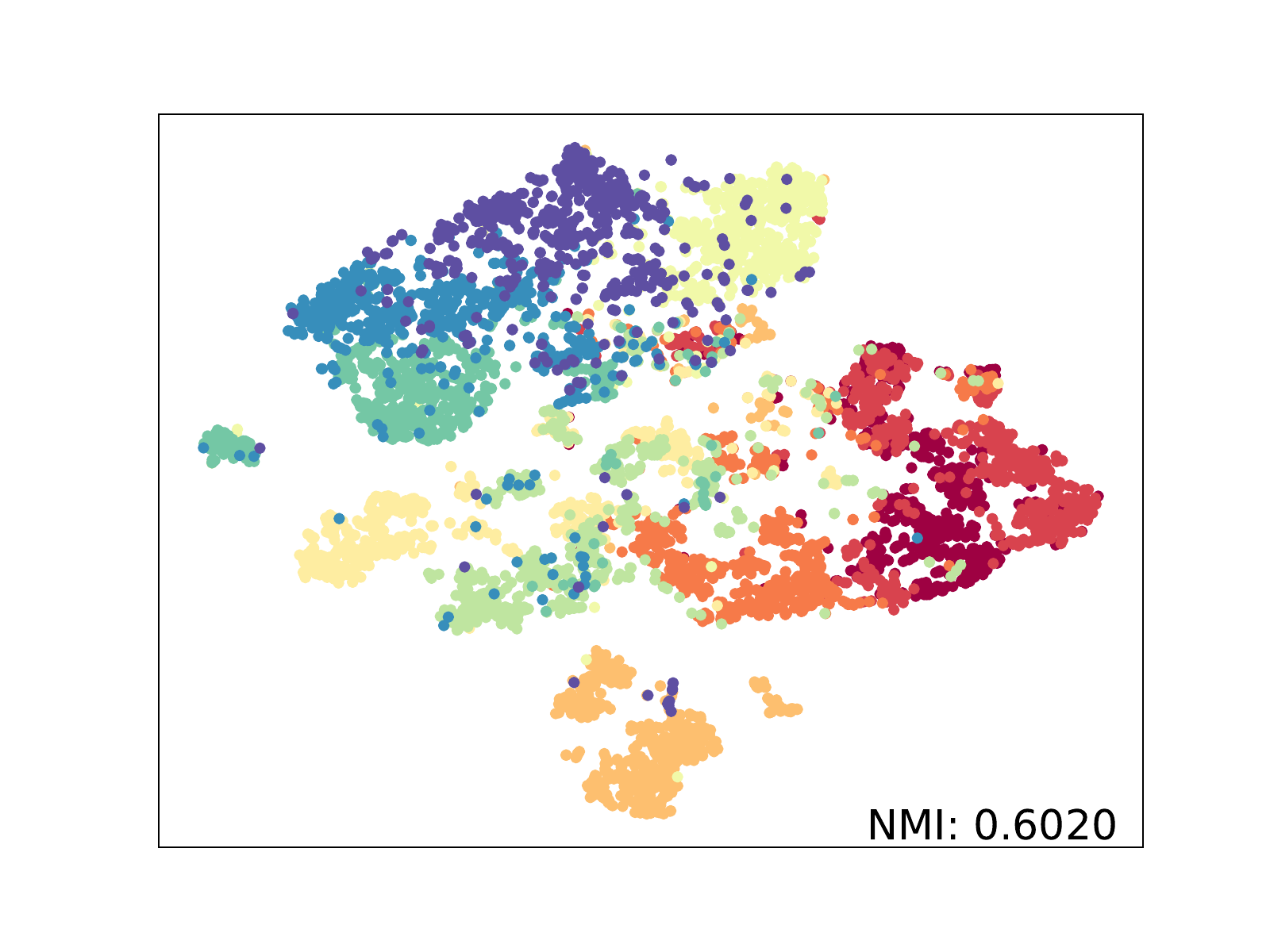}}
		\caption{The t-SNE visualization of features on NTU-60 Xsub benchmark.
		These models are trained for 300 epochs. \textbf{J}, \textbf{M}, \textbf{B} denote joint stream, motion stream, and bone stream, respectively.}
		\label{fig:tsne}
	\end{figure*}
	
	\noindent\textbf{Finetuned Evaluation Results.}
	We compare our method with other self-supervised methods and some supervised methods, such as ST-GCN, which has all the same structure and parameters as ours.
	As shown in Table~\ref{tab:finetune}, our 3s-\Method~achieves better results than supervised 3s-ST-GCN and other self-supervised methods, which indicates the effectiveness of our method.
	Moreover, in most cases, \Methodplus performs better than \Method, which indicates that the proposed multiple \aug~augmentation strategy could further boost the performance.
	
	\noindent\textbf{Semi-Supervised Evaluation Results.}
	With only 1\% and 10\% labeled data, the spatio-temporal representations learned by the model in the pretext task are important, because little data can lead to difficult convergence or overfitting problems. 
	From Table \ref{tab:semi}, our 3s-\Methodplus~outperforms other methods consistently for all configurations. 
	The results indicate that our method can make better use of spatio-temporal information, which significantly helps the model to learn better spatio-temporal representations.
	
	Based on the above experiments, we conclude that our method achieves state-of-the-art performance for self-supervised skeleton-based action recognition.
		
	\begin{table}[t]
		\caption{Finetuned results on NTU-60 and NTU-120 datasets. ``$\ddag$'' means using the bone stream data. ``$\S$'' means the model is trained with a fully-supervised manner}
		\centering
		\small
		\begin{tabular}{l|cccc}
			\toprule
			\multirow{2}{*}{Method} & \multicolumn{2}{c}{NTU-60 (\%)} & \multicolumn{2}{c}{NTU-120 (\%)} \\
			& Xsub           & Xview           & Xsub            & Xset            \\ \midrule
			\emph{Single-stream:}            &                &     & &            \\
			SkeletonCLR$^\ddag$ \cite{li20213d}    & 82.2           & 88.9            & 73.6            & 75.3            \\
			AimCLR$^{\ddag}$ \cite{guo2021contrastive}         & 83.0           & 89.2           & 77.2            & 76.0             \\
			\textbf{\Method~(Ours)}$^\ddag$   & 84.5  & 91.1   & 75.1   & 76.0   \\ 
			\textbf{\Methodplus~(Ours)}$^\ddag$   & \textbf{84.7}  & \textbf{91.8}   & \textbf{76.7}   & \textbf{78.4}   \\\midrule
			\emph{Multi-stream:}            &                &     & &            \\
			3s-ST-GCN$^\S$ \cite{yan2018spatial}     & 85.2           & 91.4            & 77.2            & 77.1            \\
			3s-CrosSCLR (\cite{li20213d})             & 86.2           & 92.5            & 80.5            & 80.4            \\
			3s-AimCLR \cite{guo2021contrastive}              & 86.9           & 92.8             & 80.1            & 80.9            \\
			\textbf{3s-\Method~(Ours)}      & \textbf{87.8}  & 93.9     & 81.6   & 81.2 \\
			\textbf{3s-\Methodplus~(Ours)}      & 87.7  & \textbf{94.0}      & \textbf{82.0}   & \textbf{82.9}\\ \bottomrule
		\end{tabular}
		\label{tab:finetune}
	\end{table}
	
	\noindent\textbf{Qualitative Results}. 
	We apply t-SNE \cite{van2008visualizing} with fixed settings to show the embeddings distribution of SkeletonCLR and our \Methodplus~on NTU-60 Xsub benchmark.
	The reported t-SNE results are fair comparisons with the same randomly selected 10 class samples and we also calculate the NMI (Normalized Mutual Information) for objective.
	The visualization results are shown in Figure~\ref{fig:tsne}.
	From the results, Our \Methodplus~always better makes the feature representation of the same class more compact and that of different classes more distinguishable.
	Furthermore, compared with SkeletonCLR, our \Methodplus~substantially improves the NMI for joint, motion, and bone streams.
	These results show that our method can extract more discriminative features for downstream tasks.
	
	\begin{table}[t]
		\caption{Semi-supervised evaluation results on PKU-MMD and NTU-60 datasets.}
		\centering
		\small
		\tabcolsep1.4mm
		\begin{tabular}{l|cc|cc}
			\toprule
			\multirow{2}{*}{Method}        & \multicolumn{2}{c|}{PKU-MMD (\%)}     & \multicolumn{2}{c}{NTU-60 (\%)} \\
			& part \uppercase\expandafter{\romannumeral1} & part \uppercase\expandafter{\romannumeral2}  & Xsub  & Xview  \\ \midrule
			\emph{1\% labeled data:}       &               &                &               &               \\
			MS$^2$L \cite{lin2020ms2l}         & 36.4          & 13.0           & 33.1          & -             \\
			ISC \cite{thoker2021skeleton}             & 37.7          & -              & 35.7          & 38.1             \\
			3s-CrosSCLR (\cite{li20213d})      & 49.7          & 10.2           & 51.1          & 50.0          \\
			3s-Colorization \cite{yang2021skeleton}  & -             & -              & 48.3          & 52.5          \\
			3s-AimCLR \cite{guo2021contrastive}       & 57.5           & 15.1           & 54.8          & 54.3           \\
			\textbf{3s-\Method~(Ours)}  & 62.2   & 15.7 & 55.3 & 55.7 \\
			\textbf{3s-\Methodplus~(Ours)}  & \textbf{62.6}    & \textbf{16.3} & \textbf{55.9} & \textbf{56.2} \\\midrule
			\emph{10\% labeled data:}      &               &                &               &               \\
			MS$^2$L \cite{lin2020ms2l}         & 70.3          & 26.1           & 65.2          & -             \\
			ISC \cite{thoker2021skeleton}             & 72.1          & -              & 65.9          & 72.5             \\
			3s-CrosSCLR (\cite{li20213d})      & 82.9          & 28.6           & 74.4          & 77.8           \\
			3s-Colorization \cite{yang2021skeleton}  & -             & -              & 71.7          & 78.9          \\
			3s-AimCLR \cite{guo2021contrastive}                   & 86.1          & 33.4           & 78.2          & 81.6         \\
			\textbf{3s-\Method~(Ours)}  & 87.7    & 41.0 & 79.9 & 83.6 \\
			\textbf{3s-\Methodplus~(Ours)}  & \textbf{88.6}    & \textbf{42.3} & \textbf{81.3} & \textbf{84.7} \\ \bottomrule
		\end{tabular}
		\label{tab:semi}
	\end{table}
	
	\section{Conclusion}
	In this paper, we propose \aug~augmentation that utilizes the topological information of skeleton data into consideration to perform spatio-temporal cropping operation, which maintains the consistency of both remaining skeleton sequences and cropped skeleton fragments, providing more informative features for contrastive learning.
	Based on \aug~augmentation strategy, we propose \Method~which uses the remaining skeleton sequences and cropped skeleton fragments to expand hard contrastive pairs, which helps the model to learn better representations.
	Extensive experiments on three datasets demonstrate the efficiency of our method and show that our method achieves state-of-the-art performance for self-supervised skeleton-based action recognition.
	Meanwhile, we also found that the information from multiple streams did not contribute significantly to our method, so we will extend \aug~to cross streams in the future.

 

\bibliographystyle{IEEEtran}
\bibliography{SkeleMixCLR}



\begin{IEEEbiography}[{\includegraphics[width=1in,height=1.25in,clip,keepaspectratio]{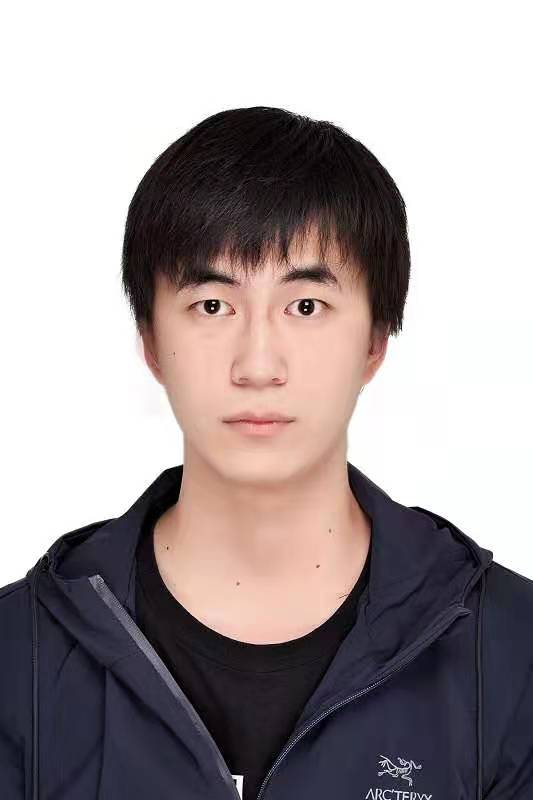}}]{Zhan Chen}
	Zhan Chen received the B.S. degree from Hunan University(HNU), China. He is a research graduate student studying at Peking University (PKU), China. His research interest lies in machine learning and computer vision.
\end{IEEEbiography}

\begin{IEEEbiography}[{\includegraphics[width=1in,height=1.25in,clip,keepaspectratio]{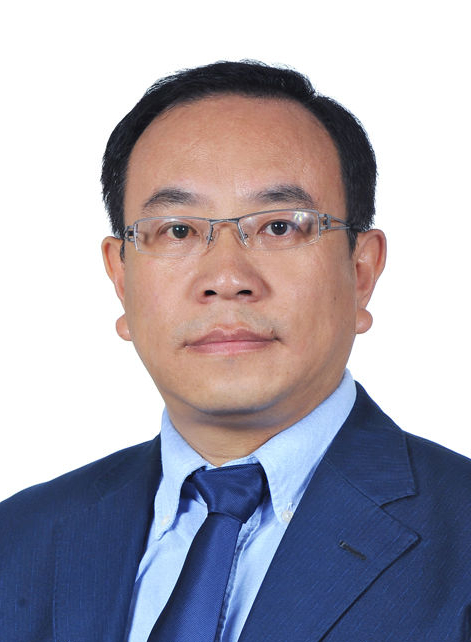}}]{Hong Liu}
	received the Ph.D. degree in mechanical electronics and automation in 1996. He serves as a Full Professor in the School of EE\&CS, Peking University (PKU), China. Prof. Liu has been selected as Chinese Innovation Leading Talent supported by National High-level Talents Special Support Plan since 2013. Dr. Liu has published more than 200 papers and gained the Chinese National Aerospace Award, Wu Wenjun Award on Artificial Intelligence, Excellence Teaching Award, and Candidates of Top Ten Outstanding Professors in PKU. He has served as keynote speakers, co-chairs, session chairs, or PC members of many important international conferences, such as IEEE/RSJ IROS, IEEE ROBIO, IEEE SMC, and IIHMSP. Recently, Dr. Liu publishes many papers on international journals and conferences, including TMM, TCSVT, TCYB, TALSP, TRO, PR, IJCAI, ICCV, CVPR, ICRA, IROS, etc.
\end{IEEEbiography}

\begin{IEEEbiography}[{\includegraphics[width=1in,height=1.25in,clip,keepaspectratio]{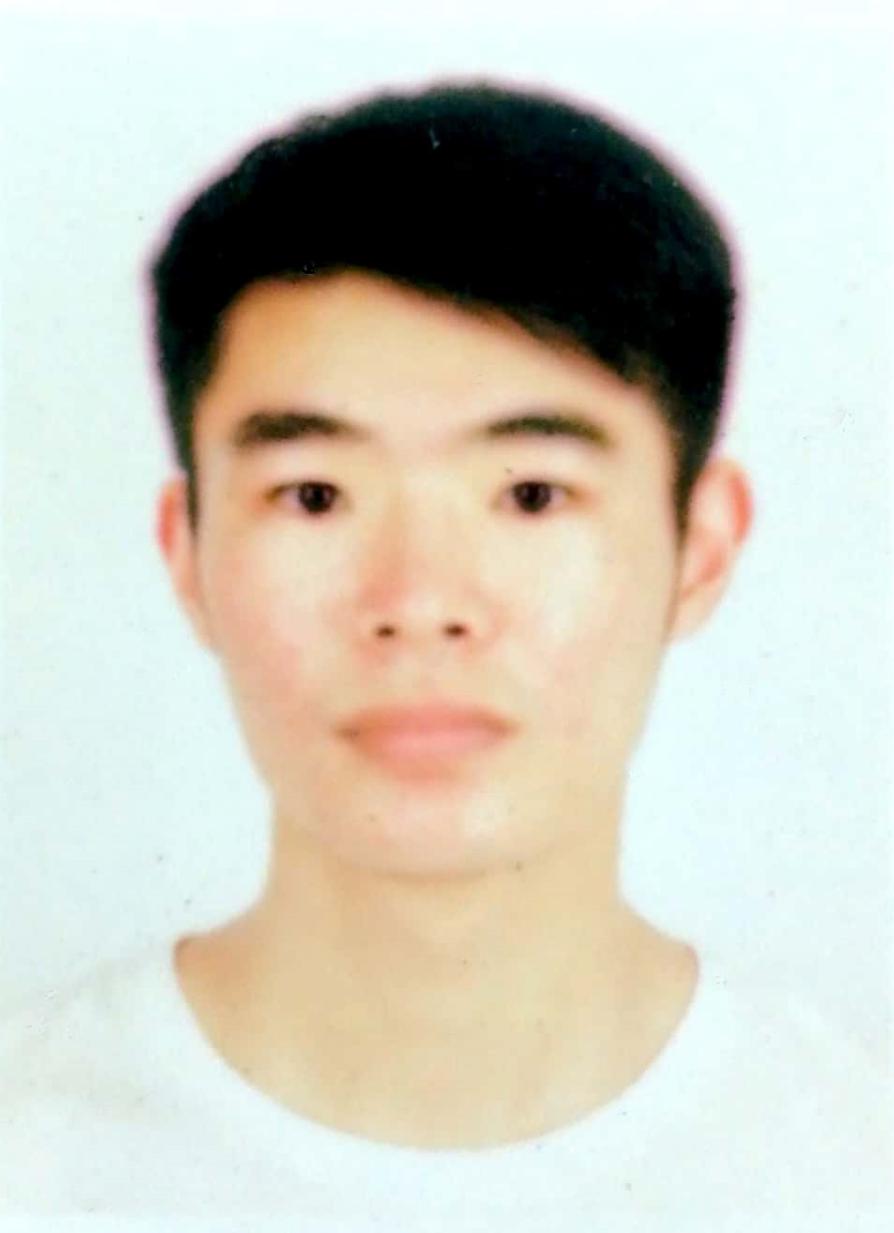}}]{Tianyu Guo} received the B.S. degree in electronics and information engineering in 2020. He is currently a graduate student in the School of Electronics and Computer Engineering, Peking University (PKU), China, under the supervision of Prof. H. Liu. His research interests include computer vision, machine learning, and action recognition.
\end{IEEEbiography}

\begin{IEEEbiography}[{\includegraphics[width=1in,height=1.25in,clip,keepaspectratio]{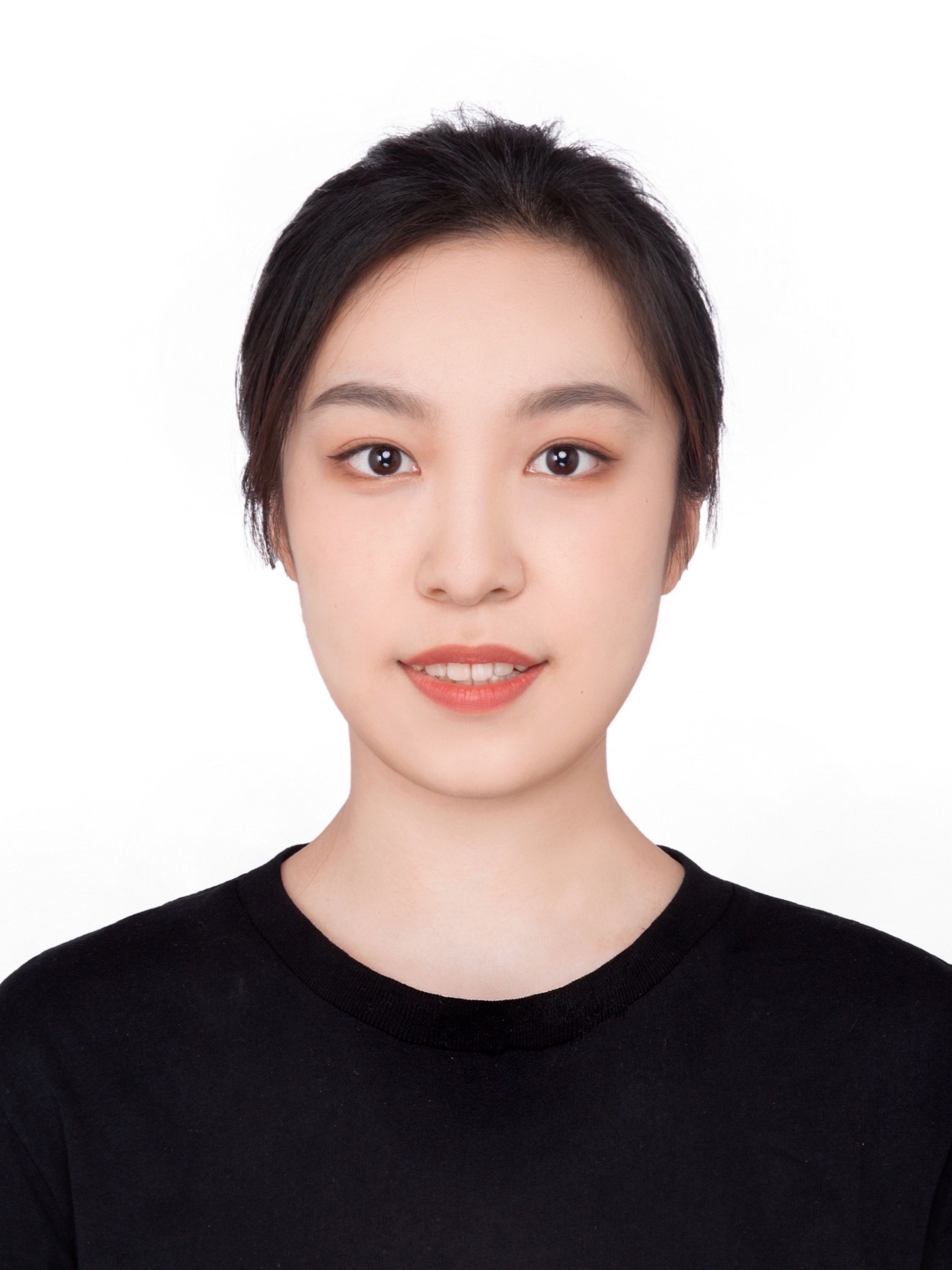}}]{Zhengyan Chen}
received the B.S. degree in information security from Hunan University, Changsha, China, in 2019. 
She is currently working toward a Master's degree in the Key Laboratory of Machine Perception, School of Electronic and Computer Engineering, Peking University (PKU), China, under the supervision of Prof. Hong Liu.
Her research interests include computer vision, human action recognition, and video analysis and understanding.
\end{IEEEbiography}

\begin{IEEEbiography}[{\includegraphics[width=1in,height=1.25in,clip,keepaspectratio]{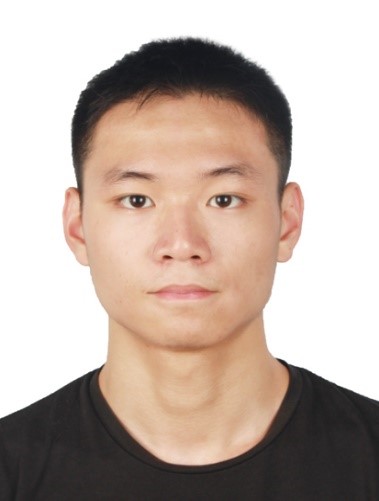}}]{Pinhao Song} received the B.E. degree in Mechanical Engineering in 2019, where he is currently pursuing the master’s degree in computer applied technology in Peking University. His current research interests include underwater object detection, generic object detection, and domain generalization.
\end{IEEEbiography}

\begin{IEEEbiography}[{\includegraphics[width=1in,height=1.25in,clip,keepaspectratio]{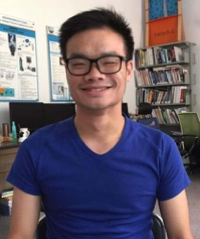}}]{Hao Tang} is currently a Postdoctoral with Computer Vision Lab, ETH Zurich, Switzerland.
He received the master’s degree from the School of Electronics and Computer Engineering, Peking University, China and the Ph.D. degree from the Multimedia and Human Understanding Group, University of Trento, Italy.
He was a visiting scholar in the Department of Engineering Science at the University of Oxford. His research interests are deep learning, machine learning, and their applications to computer vision.
\end{IEEEbiography}

\vfill

\end{document}